\PassOptionsToPackage{table,xcdraw}{xcolor}
\documentclass{article}

\usepackage{microtype}
\usepackage{graphicx}
\usepackage{subfigure}
\usepackage{booktabs} 
\usepackage{hyperref}

\usepackage[accepted]{icml2025}
\usepackage{amsmath}
\usepackage{amssymb}
\usepackage{mathtools}
\usepackage{amsthm}
\usepackage[capitalize,noabbrev]{cleveref}
\usepackage{enumitem}
\usepackage{xcolor}  
\DeclareMathOperator*{\argmax}{arg\,max}

\usepackage{multirow}
\usepackage{arydshln}
\theoremstyle{plain}

\newtheorem*{theorem*}{Theorem}

\theoremstyle{definition}

\theoremstyle{remark}

\usepackage[textsize=tiny]{todonotes}

\icmltitlerunning{Finding the Muses}

\begin{document}

\twocolumn[
\icmltitle{Finding the Muses: Identifying Coresets through Loss Trajectories}
\icmlsetsymbol{equal}{*}

\begin{icmlauthorlist}
\icmlauthor{Manish  Nagaraj}{yyy}
\icmlauthor{Deepak Ravikumar}{yyy}
\icmlauthor{Efstathia Soufleri}{yyy}
\icmlauthor{Kaushik Roy}{yyy}
\end{icmlauthorlist}

\icmlaffiliation{yyy}{Electrical and Computer Engineering, Purdue University, West Lafayette, USA}

\icmlcorrespondingauthor{Manish Nagaraj}{mnagara@purdue.edu}
\icmlkeywords{Machine Learning, ICML}

\vskip 0.3in
]

\printAffiliationsAndNotice{} 

\begin{abstract}
Deep learning models achieve state-of-the-art performance across domains but face scalability challenges in real-time or resource-constrained scenarios. 
To address this, we propose \textit{Loss Trajectory Correlation} ($\mathtt{LTC}$), a novel metric for coreset selection that identifies critical training samples driving generalization. 
$\mathtt{LTC}$ quantifies the alignment between training sample loss trajectories and validation set loss trajectories, enabling the construction of compact, representative subsets. 
Unlike traditional methods with computational and storage overheads that are infeasible to scale to large datasets, $\mathtt{LTC}$ achieves superior efficiency as it can be computed as a byproduct of training.
Our results on CIFAR-100 and ImageNet-1k show that $\mathtt{LTC}$ consistently achieves accuracy on par with or surpassing state-of-the-art coreset selection methods, with any differences remaining under $1\%$.
$\mathtt{LTC}$ also effectively transfers across various architectures, including ResNet, VGG, DenseNet, and Swin Transformer, with minimal performance degradation ($<2\%$). 
Additionally, $\mathtt{LTC}$ offers insights into training dynamics, such as identifying aligned and conflicting sample behaviors, at a fraction of the computational cost of traditional methods. 
This framework paves the way for scalable coreset selection and efficient dataset optimization.
\end{abstract}

\section{Introduction}
\label{sec:Introduction}
Deep learning (DL) models rely on vast, diverse datasets to achieve state-of-the-art performance across various domains. 
However, the computational demands of training on these massive datasets often become prohibitive, especially in real-time or resource-constrained applications. 
This raises a critical question: \textit{``Which subsets of data contribute most to effective generalization?"} 
Addressing this challenge has led to the concept of \textit{coresets}, compact and representative subsets of the training data that retain the essence of the full dataset.

Coresets enable efficient training by reducing computational and storage overhead while maintaining comparable model performance. 
Their applicability spans diverse areas such as active learning~\cite{coreset_activelearning1}, neural architecture search~\cite{coreset_nas1, coreset_nas2}, dataset distillation~\cite{coreset_dc}, and continual learning~\cite{coreset_cl_1, bilevelcoresets_borsos2020}, underscoring their significance in modern deep learning workflows.

Existing methods for coreset selection often face challenges in balancing resource efficiency with effectiveness. 
Many approaches struggle to scale to large datasets, rely on heuristics that may not align with the true data distribution \cite{craig_mirzasoleiman2020, initial_beloudah2020, icarl_rebuffi2017}, or depend on resource-intensive computations such as gradient \cite{gradmatch_killamsetty2021, tracin_pruthi2020} or curvature evaluations \cite{slocurves_garg2023}. 
Moreover, methods based on bilevel optimization  \cite{glister_killamsetty2021, refinedcoreset_xia2024, bilevelcoresets_borsos2020} are often computationally prohibitive due to the intricacy of their underlying formulations.
These constraints make them impractical for large-scale or resource-constrained applications, limiting their utility in real-world deep-learning workflows.

To address these limitations, we propose a scalable and resource-efficient method for identifying critical training samples that enhance generalization. 
Our approach introduces a novel metric, \textit{\textbf{Loss Trajectory Correlation}} ($\mathtt{LTC}$), which quantifies the alignment between the loss trajectories of a training sample and those of query samples in the validation set. 
Since the validation set is drawn from the same distribution as the test data, it serves as a proxy for generalization. 
By prioritizing training samples with high $\mathtt{LTC}$ values, we identify subsets of data that drive effective generalization.
 
Our method achieves computational efficiency through two mechanisms. 
First, loss trajectories of training samples are inherently generated during the training process, requiring no additional computation. 
Second, the intrinsic features captured by $\mathtt{LTC}$ is a property of the dataset and is independent of the learning mechanism, hence it is highly transferable across architectures. 
This transferability allows us to use smaller, efficient models to establish relationships among samples, enabling the identification of coresets that are representative of the full dataset and suitable for training larger, more complex models. 
By combining efficiency, scalability, and the ability to identify impactful data subsets, our approach enhances model performance while offering valuable insights into the structure of training data.

In this work, we make the following key contributions:
\begin{enumerate}
    \item \textbf{\textit{Loss Trajectory Correlation for Generalization:}} We introduce $\mathtt{LTC}$, a novel metric to identify critical training samples by quantifying the alignment of their loss trajectories with those of a validation set. By prioritizing samples with high positive $\mathtt{LTC}$, our method effectively identifies subsets of data that drive generalization.

    \item \textbf{\textit{Efficient and Transferable Coreset Selection:}} $\mathtt{LTC}$ enables the construction of compact, representative coresets that retain model performance while reducing computational demands. Across CIFAR-100 and ImageNet-1k, $\mathtt{LTC}$ consistently delivers competitive accuracy, matching or outperforming state-of-the-art methods for most coreset sizes. Even in cases where $\mathtt{LTC}$ does not achieve the highest accuracy, its performance remains within $1\%$ of the best method. We demonstrate that $\mathtt{LTC}$-based coresets are transferable across architectures, including ResNet, VGG, DenseNet, and Swin Transformer, with minimal performance drops ($<2\%$).

    \item \textbf{\textit{Computational and Storage Efficiency:}} $\mathtt{LTC}$ achieves exceptional scalability by utilizing the loss trajectories of training samples, which are inherently generated during the training process, and only computing the loss trajectories of query samples during evaluation. This results in a computational overhead limited to additional forward passes for the query samples, significantly fewer in number than the training samples, across all epochs. Its storage overhead scales linearly with the product of the dataset size and training epochs. In contrast, competitive methods, such as those relying on pairwise similarity~\cite{graphcut_iyer2021}, feature space distances~\cite{cal_margatina2021}, or repeated gradient computations~\cite{slocurves_garg2023}, incur substantially higher costs due to their reliance on expensive operations for all training samples. By minimizing these overheads, $\mathtt{LTC}$ effectively balances scalability and efficiency, making it an optimal choice for large-scale datasets and resource-constrained applications.
    
    \item \textbf{\textit{Insights into Training Dynamics:}} We empirically show that $\mathtt{LTC}$ provides insights similar to traditional Training Data Attribution (TDA) metrics, such as identifying aligned and conflicting sample dynamics, but at significantly reduced computational cost. 
    This insight into the link between TDA and $\mathtt{LTC}$ establishes $\mathtt{LTC}$ as an excellent choice for coreset selection.
\end{enumerate}

\section{Related Literature}
Existing coreset selection techniques can be broadly categorized into score-based, optimization-based, and training property-based methods. 

\textbf{\textit{Score-based}} methods select training samples based on predefined metrics in the feature space, but they often rely on heuristics that may not reflect the true data distribution.
Common approaches include selecting samples based on their proximity to class centers or decision boundaries \cite{icarl_rebuffi2017,endtoend_castro2018,initial_beloudah2020}.
Examples of such methods include $\mathtt{DeepFool}$~\cite{deepfool_mossavi2016} and $\mathtt{BoundarySetCCS}$~\cite{mindboundary_yang2024}.
Other methods leverage the model predictions. 
$\mathbb{D}^2\mathtt{Pruning}$~\cite{modelpred_d2_maharana2023} utilizes the difficulty of prediction as a metric. 
In contrast, \cite{uncertainity_he2024} utilizes the uncertainty of prediction, and $\mathtt{E2LN}$~\cite{E2LN_grand_paul2021} measures the $L_{2}$ norm of the prediction error.
While these techniques offer valuable insights, they face scalability issues with large datasets and risk introducing biases toward specific data subsets. 
To address these limitations, methods like 
$\mathtt{GraNd}$~\cite{E2LN_grand_paul2021}, $\mathtt{GradMatch}$~\cite{gradmatch_killamsetty2021}, $\mathtt{Forgetting}$~\cite{forgetting_toneva2018}, and $\mathtt{CRAIG}$~\cite{craig_mirzasoleiman2020} incorporate network-driven metrics, such as gradient norms.
However, these approaches often depend heavily on fixed metrics, reducing their adaptability throughout the training process.

\textbf{\textit{Optimization-based}} methods, such as $\mathtt{Glister}$~\cite{glister_killamsetty2021}, $\mathtt{LBCS}$~\cite{refinedcoreset_xia2024}, and \cite{bilevelcoresets_borsos2020}, are grounded in robust theoretical frameworks but involve computationally expensive formulations, such as bilevel optimization, limiting their feasibility for large-scale datasets.

\textbf{\textit{Training property-based}} methods identify influential samples by examining their impact on generalization. 
Techniques such as $\mathtt{Slocurv}$~\cite{slocurves_garg2023} utilize second-order loss statistics to identify samples better suited for generalization, while $\mathtt{TracIn}$~\cite{tracin_pruthi2020} tracks gradient alignment between training samples and a validation set. 
Despite their effectiveness, these methods often rely on computationally intensive first-order (gradients) or second-order (curvature, Hessian) metrics, limiting their scalability for large datasets.

By introducing $\mathtt{LTC}$, our method addresses these limitations by offering a computationally lightweight yet effective alternative for coreset selection. 
It is designed to balance efficiency and effectiveness, making it applicable across a range of deep learning applications.

\section{Loss Trajectory Correlation ($\mathtt{LTC}$)}

\subsection{Overview}
\label{sec:LTC}
\begin{figure}[t]
    \centering
    \includegraphics[width=1\linewidth]{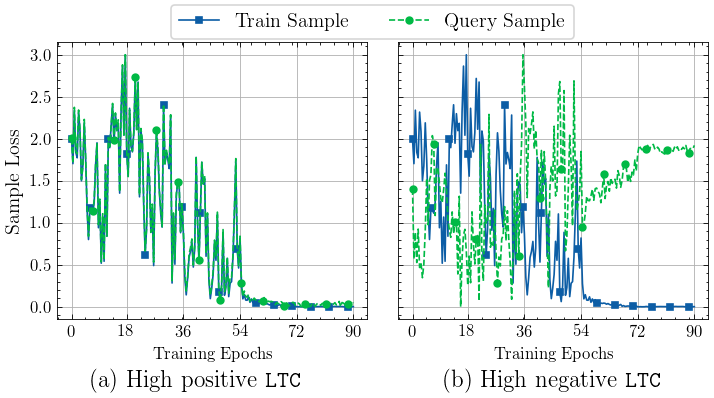}
    \vspace{-7mm}
    \caption{Examples of loss trajectories of train and query sample pairs with high positive and high negative $\mathtt{LTC}$: (\textbf{a}) shows high positive $\mathtt{LTC}$, where train (solid blue) and query (dashed green) sample losses decrease together, indicating aligned learning dynamics. (\textbf{b}) shows high negative $\mathtt{LTC}$, where reductions in the train sample’s loss correspond to increases in the query sample’s loss, highlighting conflicting relationships. These examples demonstrate how $\mathtt{LTC}$ captures inter-sample influences during training. (Best viewed in color.)}
    \label{fig:ltc_intuition}
\end{figure}

We begin by establishing key notations used throughout this paper for clarity and consistency. 
Random variables will be represented in bold ($\mathbf{V}$), with scalar instances denoted by lowercase letters ($v$), and vectors by arrowed letters ($\vec{v}$).

Consider a randomized learning algorithm $\xi$ (e.g., SGD, Adam) training over $T$ epochs with training parameters $\lambda$ on a dataset $S \sim \mathbf{Z}$, $S =\{\vec{z}_1,\ldots,\vec{z}_N\}$, where $\vec{z}_m$ is the $m^{\text{th}}$ sample. 
Let $\theta^t_S \in \mathbb{R}^p$ be the model parameters at epoch $t$, and the loss function $\ell(\theta^t_S,\vec{z})$ and its gradient $\nabla_\theta \ell(\theta^t_S,\vec{z})$ be:
$$
\ell(\theta^t_S,\vec{z}): \mathbb{R}^p \times \mathcal{Z} \to \mathbb{R}, \quad \nabla_\theta \ell(\theta^t_S,\vec{z}) \in \mathbb{R}^p
$$

For a sample $\vec{z}$, the loss change between epochs $t-1$ and $t$ is:
$$\Delta \ell(\theta^t_S,\vec{z}) = \ell(\theta^t_S, \vec{z}) - \ell(\theta^{t-1}_S, \vec{z})$$

Given a training sample \(\vec{z}_m \in S\) and a query (test or validation) sample \(\vec{z}_q \notin S\), the \(\mathtt{LTC}\) score is defined as the correlation ($\rho$) between their loss change trajectories:
\begin{align}
\mathtt{LTC}(\vec{z}_q, \vec{z}_m, S, \lambda) \coloneqq 
    \rho \bigg( 
        \big\{ \Delta \ell\big(\theta^t_S, \vec{z}_m\big) \big\}_{t=1}^T, & \notag \\
        \big\{ \Delta \ell\big(\theta^t_S, \vec{z}_q\big) \big\}_{t=1}^T 
    \bigg) 
\end{align}
For our experiments, we chose to use Pearson's correlation~\cite{pearsoncorr} due to its scale invariance and ease of computation. 

\subsection{Intuition Behind the $\mathtt{LTC}$ Metric}

The Loss Trajectory Correlation ($\mathtt{LTC}$) metric offers a powerful lens through which to understand how a model's learning dynamics for one sample influence its performance on another. 
It quantifies the relationship between the changes in loss for a training sample $\vec{z}_m$ and a query sample $\vec{z}_q$ across training. 
At its core, $\mathtt{LTC}$ measures the degree to which the loss trajectories of these two samples align or diverge during the learning process.

\paragraph{Positive $\mathtt{LTC}$: Aligned Learning Dynamics} A positive $\mathtt{LTC}$ value indicates that the loss trajectories for $\vec{z}_m$ and $\vec{z}_q$ exhibit synchronized behavior (shown in \cref{fig:ltc_intuition} (a)). 
Specifically, if the model's loss on $\vec{z}_m$ decreases at a given epoch, the loss on $\vec{z}_q$ tends to decrease as well, and vice versa. 
This alignment suggests that the features of $\vec{z}_m$ are highly relevant to predicting $\vec{z}_q$, and learning from $\vec{z}_m$ enhances the model's performance on $\vec{z}_q$. 
Such positive correlations often arise when the two samples share similar patterns, structures, or features, and the model leverages knowledge gained from $\vec{z}_m$ to infer on $\vec{z}_q$ better.

\paragraph{Negative $\mathtt{LTC}$: Conflicting Learning Dynamics} Conversely, a negative $\mathtt{LTC}$ value reveals opposing loss trajectories (shown in \cref{fig:ltc_intuition} (b)). 
In this case, reducing the loss on $\vec{z}_m$ increases the loss on $\vec{z}_q$, indicating conflicting learning dynamics. 
This phenomenon may occur when $\vec{z}_m$ and $\vec{z}_q$ encode dissimilar or contradictory features, causing the model to prioritize one at the expense of the other. 
Such conflicts provide insight into potential trade-offs in the learning process and highlight the diverse interactions between samples during training.

\begin{figure}[t]
    \centering
    \includegraphics[width=1\linewidth]{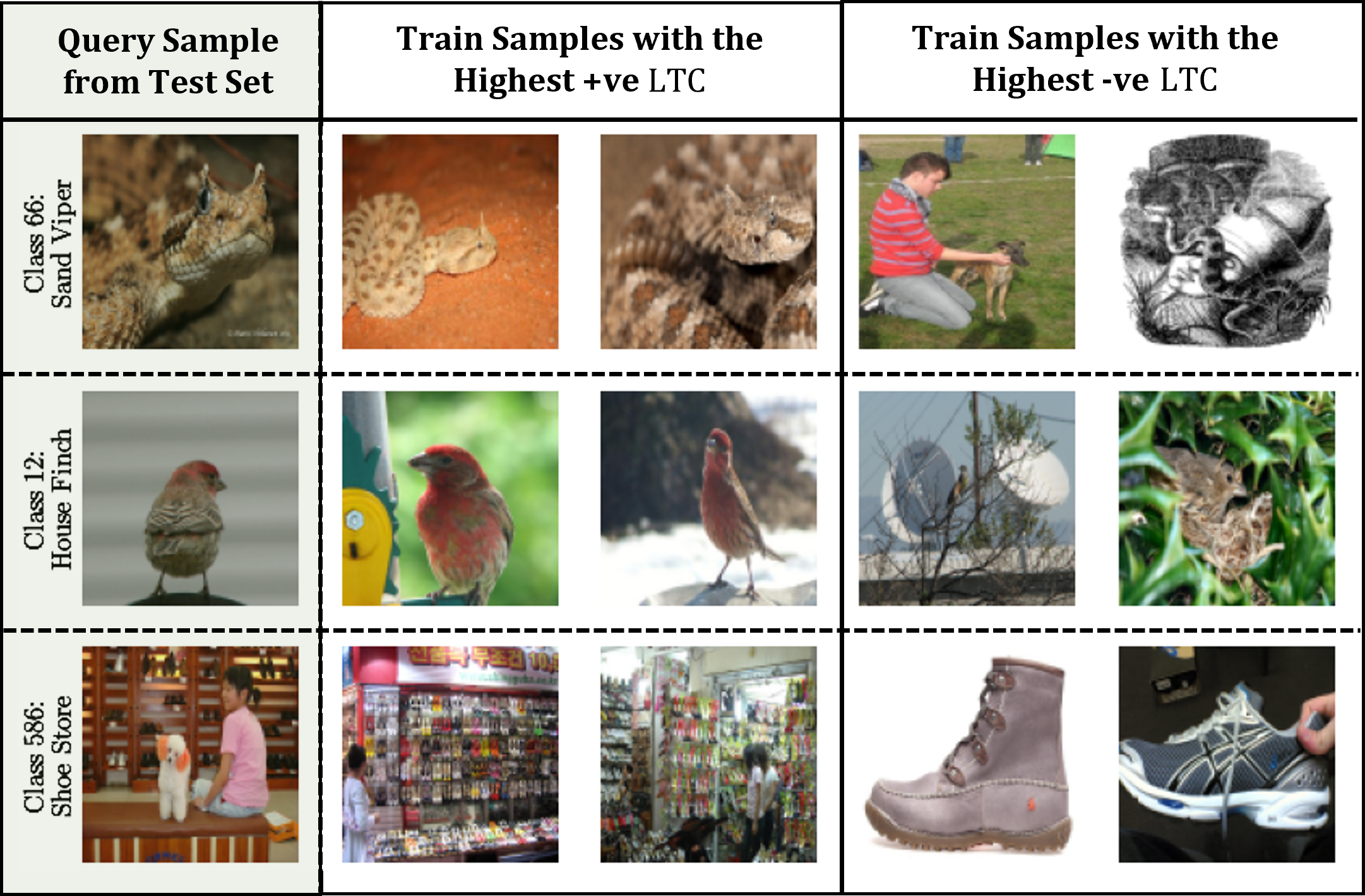}
    \vspace{-7mm}
    \caption{Examples of randomly chosen query samples from the test set in ImageNet-1k and their corresponding training samples within the same class with the highest positive $\mathtt{LTC}$ (closely aligned learning dynamics) and highest negative $\mathtt{LTC}$ (conflicting learning dynamics). This visualization highlights the alignment and contrast between query samples and training samples in terms of their feature relevance during the learning process. The $\mathtt{LTC}$ values were calculated using ResNet-18. }
    \label{fig:ltc_most_least_imagenet}
\end{figure}

\paragraph{Why $\mathtt{LTC}$ Matters} The intuitive patterns captured by $\mathtt{LTC}$, a positive correlation for aligned samples and a negative correlation for conflicting ones, make it a valuable tool for analyzing the inter-sample influences within a dataset. 
By examining the $\mathtt{LTC}$ scores between a validation set and training samples, researchers can identify which training samples contribute most effectively to generalization and which introduce conflicting signals. 
This is illustrated in \cref{fig:ltc_most_least_imagenet}, where randomly chosen query samples in ImageNet-1k and their most influential training samples (with the highest positive and negative $\mathtt{LTC}$) within the same class are shown.
This insight can be leveraged to construct influential coresets, enabling more efficient training by retaining only the most impactful samples.

In summary, $\mathtt{LTC}$ provides a principled way to probe and interpret the relationships between samples in the context of model learning. 
It allows practitioners to trace how individual samples affect others during training and offers actionable insights into dataset optimization and model behavior.

\section{Coreset Identification}
\label{sec:coreset_identification}
\begin{figure*}[t]
    \centering
    \includegraphics[width=1\linewidth]{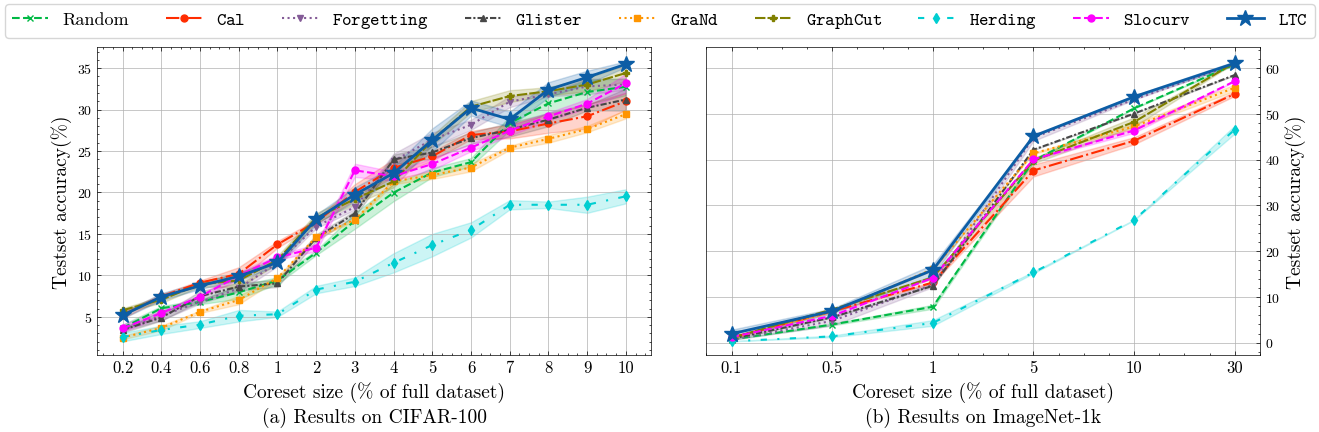}
    \vspace{-7mm}
    \caption{Comparison of test accuracy across various coreset selection methods on the CIFAR-100 and ImageNet-1k datasets. The proposed approach consistently outperforms or matches existing techniques for all evaluated dataset sizes, demonstrating its effectiveness. The shaded regions in the plots represent the standard deviation across five random seeds. (Best viewed in color.)}
    \label{fig:coreset_baselines}
    \vspace{-3mm}
\end{figure*}

\subsection{Methodology}
\label{sec:coreset_methodology}
The $\mathtt{LTC}$ metric offers a systematic approach to identifying impactful training samples by quantifying their influence on the validation set through an analysis of loss trajectories. 

As detailed in \cref{alg:coreset}, a source model $\theta$ is trained on the dataset $S$ while recording the loss trajectories for both training and validation ($\mathcal{V}$) samples. 
The $\mathtt{LTC}$ score is calculated as the correlation between the loss trajectories of each training sample and each validation sample. 
To rank and select impactful samples, the average $\mathtt{LTC}$ score for each training sample is computed, allowing the top-$k$ samples to be identified and used to form the coreset. 

A key advantage of this approach lies in its flexibility, the source model used to compute $\mathtt{LTC}$ does not need to match the model for which the coreset will ultimately be applied. 
This enables the method to be employed across diverse architectures and tasks, enhancing its practical utility.
\renewcommand\algorithmicrequire{\textbf{Input:}}
\renewcommand\algorithmicensure{\textbf{Output:}}
\begin{algorithm}[ht]
\small
\caption{Coreset Selection Using $\mathtt{LTC}$}
\label{alg:coreset}
\begin{algorithmic}[1]
\REQUIRE  Training set $S = \{\vec{z}_m\}_{m=1}^N$, Validation set $\mathcal{V} = \{\vec{z}_q\}_{q=1}^Q$, Coreset size $k$, Number of epochs $T$, Loss function $\ell$, Randomized Learning Algorithm $\xi$, Training parameters $\lambda$, source model $\theta$
\ENSURE Coreset $\mathcal{C}$ of size $k$

\vspace{0.2cm}

\STATE \textcolor{gray}{\textit{// Train Model and Record Loss Trajectories}}
\FOR{epoch $t = 1, \ldots, T$}
    \STATE $\theta^t_S \gets \xi(\theta^{t-1}_S, S, \lambda)$ 
    \quad $\triangleright$ \textit{Update model weights}
    \FOR{each sample $\vec{z}_m \in S$}
        \STATE Compute and store $\ell(\theta^t_S, \vec{z}_m)$
    \ENDFOR
    \FOR{each sample $\vec{z}_q \in \mathcal{V}$}
        \STATE Compute and store $\ell(\theta^t_S, \vec{z}_q)$
    \ENDFOR
\ENDFOR

\vspace{0.2cm}
\STATE \textcolor{gray}{\textit{// Calculate $\mathtt{LTC}$ Scores}}
\STATE Initialize $\mathtt{LTC}(q, m) \gets 0, \forall m \in S, q \in \mathcal{V}$
\FOR{each $\vec{z}_m \in S$ and $\vec{z}_q \in \mathcal{V}$}
    \STATE $\mathtt{LTC}(q, m) \gets \rho\big(\{\Delta \ell(\theta^t_S, \vec{z}_m)\}_{t=1}^T,$
    \STATE \hspace{7.5em} $\{\Delta \ell(\theta^t_S, \vec{z}_q)\}_{t=1}^T \big)$
\ENDFOR

\vspace{0.2cm}
\STATE \textcolor{gray}{\textit{// Aggregate and Rank Training Samples}}
\FOR{each $\vec{z}_m \in S$}
    \STATE $\mathtt{LTC}_{\text{Avg}}(m) \gets \frac{1}{Q} \sum_{q=1}^Q \mathtt{LTC}(q, m)$
\ENDFOR
\STATE $\triangleright$ Rank training samples $\{\vec{z}_m\}_{m=1}^N$ by $\mathtt{LTC}_{\mathrm{Avg}}(m)$ 

\vspace{0.2cm}
\STATE \textcolor{gray}{\textit{// Select Top \(k\) Samples}}
\STATE $\mathcal{C} \gets$ \text{Top \(k\) samples from \(S\) by the above ranking}

\vspace{0.2cm}
\RETURN $\mathcal{C}$

\end{algorithmic}
\end{algorithm}
\vspace{-3mm}

\subsection{Experimental Setup}
\label{sec:coresets_experiments}
To evaluate the effectiveness of the proposed metric in identifying coresets, we conducted experiments on two widely-used datasets: CIFAR-100~\citep{cifar} and ImageNet-1k~\citep{imagenet}. 
CIFAR-100 contains $50,000$ training samples and $10,000$ test samples distributed across $100$ classes, while ImageNet-1k is a significantly larger dataset with $1,281,167$ training samples and $50,000$ validation samples across $1000$ classes. 
Unless stated otherwise, all experiments used the ResNet-18~\citep{resnet} architecture.

We evaluated coresets of varying sizes: $0.2\%$ to $10\%$ of the full dataset for CIFAR-100 and $0.1\%$ to $30\%$ for ImageNet-1k. 
Each coreset was constructed to maintain class balance, ensuring equal representation across all classes. 
The training configuration, including augmentation strategies and optimizer details, is detailed in \cref{appendix:coreset_gen}.

\subsection{Comparison to Baseline Methods}
\label{sec:baseline_comparison_coreset}
We compared our approach, $\mathtt{LTC}$, to several state-of-the-art coreset selection techniques, including $\mathtt{Glister}$~\citep{glister_killamsetty2021}, $\mathtt{Forgetting}$~\citep{forgetting_toneva2018}, $\mathtt{GraphCut}$~\citep{graphcut_iyer2021}, $\mathtt{Cal}$~\citep{cal_margatina2021}, $\mathtt{GraNd}$~\citep{E2LN_grand_paul2021}, $\mathtt{Herding}$~\citep{herding_chen2010}, and $\mathtt{Slocurv}$~\citep{slocurves_garg2023}. 
Implementations of these methods were sourced from the DeepCore library~\citep{deepcore}. 
To ensure uniformity, all methods were evaluated under the same experimental conditions, including identical training and testing setups. 
For coreset generation methods requiring pretraining, models were trained for 40 epochs to achieve convergence under identical conditions across all techniques. 
No additional fine-tuning or regularization was applied to ensure fair comparisons.

\textbf{Results and Observations:} As shown in \cref{fig:coreset_baselines}, $\mathtt{LTC}$ demonstrated exceptional performance across a wide range of coreset sizes. 
On CIFAR-100, $\mathtt{LTC}$ consistently outperformed or matched the best-performing baselines, particularly for moderate to larger coresets (e.g., $5\%$--$10\%$ of the dataset). 
At $5\%$, $\mathtt{LTC}$ achieved a mean accuracy of $26.33\%$, surpassing both $\mathtt{Cal}$ and $\mathtt{GraphCut}$. 
At $10\%$, $\mathtt{LTC}$ reached $35.47\%$, exceeding the next best performers $\mathtt{GraphCut}$ by $1.06\%$ and $\mathtt{SloCurv}$ by $2.30\%$. 
For smaller coresets, $\mathtt{LTC}$ remained competitive, with performance differences of less than $1\%$ compared to the leading methods $\mathtt{Cal}$ and $\mathtt{GraphCut}$.

On ImageNet-1k, $\mathtt{LTC}$ maintained a leading or near-leading position across all coreset sizes. 
At even small subsets (e.g., $0.1\%$, it achieved $1.95\%$ accuracy. 
For medium coresets (e.g., $5\%$--$10\%$), $\mathtt{LTC}$ achieved $45.15\%$ and $53.78\%$, outperforming most methods except $\mathtt{SloCurv}$ at $5\%$. 
Even for large coresets (e.g., $30\%$), $\mathtt{LTC}$ remained competitive, achieving $61.11\%$, only $0.13\%$ behind $\mathtt{GraphCut}$. 

For a detailed numerical breakdown of the results, please refer to \cref{appendix:coreset_gen}, where the same findings are presented in tabular format for clarity and precision.

\textbf{Takeaways:} The experimental results highlight the robustness and effectiveness of $\mathtt{LTC}$ across both CIFAR-100 and ImageNet-1k datasets. 
It consistently ranked as the top or near-top method across all evaluated coreset sizes.
This adaptability and superior performance across varying dataset scales and coreset sizes underscore the versatility of $\mathtt{LTC}$ in identifying representative subsets for diverse applications.

\begin{figure*}[t]
    \centering
    \includegraphics[width=1\linewidth]{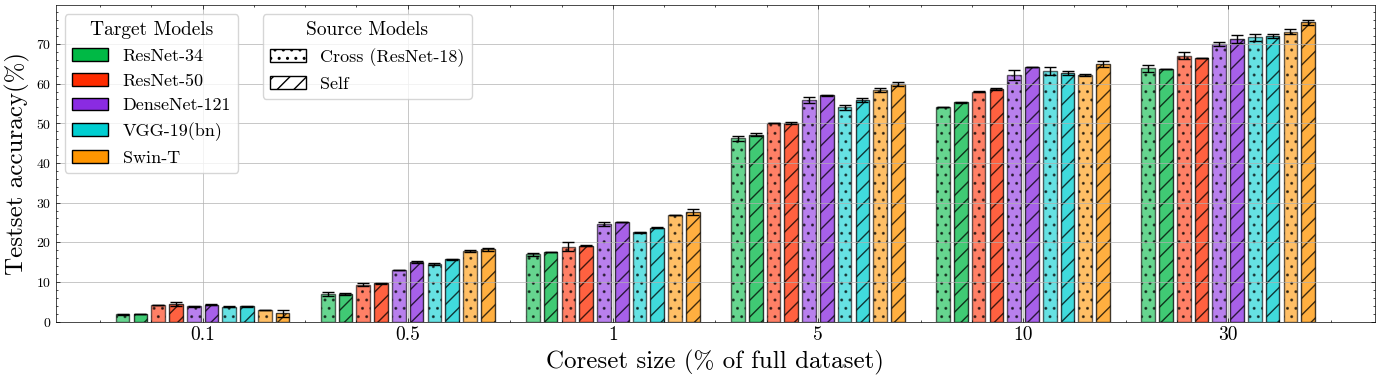}
    \vspace{-7mm}
    \caption{Comparison of the performance of coresets of different sizes of ImageNet-1k on different target models (shown in different colors) identified by the same architecture as the source model (shown as striped bars) to the coresets identified by a different (smaller) source model, ResNet-18 (shown as dotted bars). The error bars show the variance of the performance over 5 runs. We see a very minimal drop in accuracy when the smaller source models is used. (Best viewed in color.)}
    \vspace{-3mm}
    \label{fig:coreset_transferability}
\end{figure*}
\subsection{Transferability across Architectures}
\label{sec:transferability}
As discussed in \cref{sec:coreset_methodology}, the coresets are identified using a source model $\theta$. 
An advantage of using $\mathtt{LTC}$ becomes evident when we can effectively use a smaller, more efficient source model to identify coresets for training a larger, more complex target model. 
This is especially useful in scenarios where the target model has significantly more parameters and requires greater computational resources for training.

The rationale behind this approach stems from the observation that, irrespective of the specific architectures employed (e.g., deep neural networks or vision transformers), data points with similar characteristics exhibit similar feature representations. 
Consequently, the $\mathtt{LTC}$ scores, which quantify the relative importance of data points, should remain consistent across different architectures. 
This hypothesis aligns with the inherent property of feature similarity and serves as the foundation for our empirical studies.

To validate this hypothesis, we conducted experiments where we used ResNet-18 as the source model to compute coresets of varying sizes on the ImageNet-1k dataset. 
These coresets were then utilized to train diverse target architectures. The target architectures included ResNet-34 and ResNet-50, VGG-19 with batch normalization (referred to as VGG-19(bn)), DenseNet-121, and Swin-T Transformer.
The results using the ResNet-18 source model were compared to results obtained when the respective target models themselves were used as source models. 
This comparative analysis allowed us to assess the impact of the source model's architecture on the effectiveness of the coresets for training various target models.

\textbf{Results and Observations:} The results presented in \cref{fig:coreset_transferability} confirm the high transferability of coresets identified by a smaller source model, to various target architectures. 
Across all evaluated target models and coreset sizes, the performance achieved using coresets identified by ResNet-18 consistently exhibited minimal accuracy drops when compared to those identified using the respective target models as source models. 
For instance, in the case of ResNet-34, the accuracy at a $10\%$ coreset size was $54.13\%$ using ResNet-18 as the source model, compared to $55.31\%$ when ResNet-34 was used as the source model, representing a negligible difference of $1.18\%$. 
Similarly, for other target models, including DenseNet-121, VGG-19(bn), and Swin-T, the accuracy differences remained within $2\%$ across all coreset sizes, even for small subsets like $0.1\%$ and $0.5\%$. 
Importantly, this trend held consistently across diverse architectures, including convolutional neural networks and vision transformers, underscoring the architecture-agnostic nature of the coresets identified using $\mathtt{LTC}$. 
Moreover, the results exhibited low variance across multiple experimental runs, reinforcing the reliability of the approach. 
These findings validate the hypothesis that coresets generated by $\mathtt{LTC}$ retain their effectiveness across different target architectures, regardless of the source model used to identify them.

For a detailed numerical breakdown of the results, please refer to \cref{appendix:coreset_gen}, where the same findings are presented in tabular format for clarity and precision.

\textbf{Takeaways:} The experimental results demonstrate that coresets identified by $\mathtt{LTC}$ using a smaller source model, such as ResNet-18, are highly transferable and exhibit minimal performance degradation when used to train larger and more complex target architectures. 
This not only highlights the robustness of $\mathtt{LTC}$ but also emphasizes its practical utility in enabling resource-efficient coreset identification. 

\subsection{Computational and Storage Overheads} 
\label{sec:coreset_overheads}
The computational overhead of $\mathtt{LTC}$ is minimal, as the loss trajectories of training samples are inherently generated during the training process. 
For query samples, their loss trajectories are computed during evaluation, requiring only forward passes across epochs. 
In contrast, competitive methods like $\mathtt{Cal}$ and $\mathtt{GraphCut}$ involve costly nearest-neighbor searches and pairwise similarity computations, while $\mathtt{Slocurv}$ relies on repeated gradient-based calculations. 
\cref{tab:coreset_overheads} summarizes the overheads of these methods, with detailed derivations in \cref{appendix:details_overheads_coresets}.
\begin{table}[t]
\centering
\vspace{-3mm}
\caption{Comparison of computational and storage overheads of $\mathtt{LTC}$ with existing coreset generation methods. The details of the notation are outlined in \cref{sec:coreset_overheads}.}
\vspace{2mm}
\label{tab:coreset_overheads}
\renewcommand{\arraystretch}{1.25}
\begin{tabular}{|c|c|c|}
\hline
\rowcolor[HTML]{D8D8D8} \textbf{Method} & \begin{tabular}[c]{@{}c@{}} \textbf{Computational}\\\textbf{Overhead} \end{tabular}& \begin{tabular}[c]{@{}c@{}} \textbf{Storage}\\\textbf{Overhead} \end{tabular} \\\hline \hline
{\small$\mathtt{Glister}$} & $O(\frac{NQTf}{\gamma}\log (1/\epsilon))$ & $O(Q)$ \\
{\small$\mathtt{Forgetting}$} & - & $O(NT)$ \\
{\small$\mathtt{GraphCut}$} & $O(N^2k)$ & $O(N^2)$ \\
{\small$\mathtt{Cal}$} & $O(NQd)$ & $O(Nd)$ \\
{\small$\mathtt{GraNd}$} & $O(3NTRf)$ & $O(NTRp)$ \\
{\small$\mathtt{Herding}$} & $O(NTd)$ & $O(Nd)$ \\
{\small$\mathtt{Slocurv}$} & $O(3NRf)$ & $O(NRd)$ \\\hdashline
\rowcolor[HTML]{E6E6FF} $\mathtt{LTC}$ (Ours) & $O(QTf)$ & $O(NT)$ \\\hline
\end{tabular}
\vspace{-5mm}
\end{table}

\textbf{Notation and Setup:} Let $N$ and $Q$ denote the number of training and query samples, respectively, and $T$ the number of training epochs. 
Computational complexities are expressed in terms of \textit{floating-point operations per second} (FLOPs) proportional to these variables. 
The storage overheads reflect the additional disk space required for dataset and model parameters ($p$). 
Other parameters include $k$, the coreset size; $\gamma$, the frequency of coreset generation (used in $\mathtt{Glister}$); $d$, the input dimensionality; $c$, the number of classes; and $R$, the number of retrainings required for $\mathtt{GraNd}$ and $\mathtt{Slocurv}$. 
Additionally, $f$ refers to the FLOPs required for a single forward pass through the model, and, a backward pass is approximately twice as expensive as a forward pass and involves $2f$ FLOPs.

\textbf{Results and Observations:} $\mathtt{LTC}$ demonstrates significantly lower computational and storage overheads compared to competitive methods. 
Specifically, $\mathtt{GraphCut}$ requires $O(N^2k)$ operations due to pairwise similarity evaluations, while $\mathtt{Cal}$ incurs $O(NQd)$ operations for feature space distance computations, and $\mathtt{Slocurv}$ scales as $O(3NRf)$ because of repeated gradient evaluations. 
In contrast, $\mathtt{LTC}$ only scales as $O(QTf)$. 
Due to the fact that $N \gg Q$ in large datasets and $T$ is typically small, $\mathtt{LTC}$ incurs significantly lower overheads compared to competing methods.
For instance, on ImageNet-1k with ResNet-18, identifying a coreset of $10\%$ size, $\mathtt{LTC}$ incurs computational and storage overheads of around $8$ PFLOPs\footnote{peta FLOPs ($10^{15}$ FLOPs)} and $0.4$ GB\footnote{assuming each parameter needs 4B of storage space}, respectively. 
In comparison, $\mathtt{GraphCut}$ and $\mathtt{Slocurv}$ require around $210$ and $69$ PFLOP operations, with storage footprints of around $6500-7700$ GB.
Although $\mathtt{Cal}$ has computational overheads closer to $\mathtt{LTC}$, its storage requirements (around $771$~GB) are significantly higher. 
Detailed calculations for this example are provided in \cref{appendix:summarized_coreset_overheads}.

\textbf{Takeaways:} $\mathtt{LTC}$ provides a computationally and memory-efficient approach to coreset selection, outperforming methods like $\mathtt{GraphCut}$, $\mathtt{Slocurv}$, and $\mathtt{Cal}$ in scalability and resource efficiency. These advantages make $\mathtt{LTC}$ well-suited for large-scale and resource-constrained applications.

\section{Discussion}
\label{sec:discussion}
In this section, we delve deeper into the insights provided by $\mathtt{LTC}$ regarding training dynamics, specifically its ability to identify the impact of training samples on generalization.

\subsection{Comparison of $\mathtt{LTC}$ with Influence}
\label{sec:TDA}
The impact measured by $\mathtt{LTC}$ closely aligns with the  ``\textit{influence}" of individual training samples on a model's predictions that are measured by \textit{Training Data Attribution} (TDA) methods. 
TDA methods have been widely employed for tasks such as debugging datasets, interpreting models, and optimizing training efficiency~\cite{Koh2017, representer_yeh2018, FZ_infl_feldman2020}. 
Early approaches, like \textit{Leave-One-Out} (LOO) retraining, involved repeatedly retraining models after removing specific data points but were computationally impractical for modern deep learning. 
Recent metrics, such as FZ-Influence ($\mathtt{Infl}$)~\cite{FZ_infl_feldman2020} and $\mathtt{Datamodels}$~\cite{datamodels_ilyas2022}, introduced precomputed scores for popular datasets but struggled with scalability. 
\textit{Influence Functions} offered an alternative by estimating sample effects using gradient and Hessian computations, with efficiency improvements from methods like $\mathtt{RandSelect}$~\cite{Randselect_wojnowicz2016}, $\mathtt{Arnoldi}$ iterations~\cite{Arnoldi_schioppa2022}, and $\mathtt{TRAK}$~\cite{trak_2023park}. 
However, these approaches relied on strong assumptions and incurred high computational costs. 
\textit{Unrolling-based methods}, such as $\mathtt{TracIn}$~\cite{tracin_pruthi2020} addressed some of these issues by tracking gradients throughout training, but their need to store intermediate training states resulted in significant memory and computational overhead, limiting their practicality for large-scale applications. 
A detailed overview of TDA metrics is provided in \cref{appendix:TDA_metrics_overview}.

In contrast, $\mathtt{LTC}$ solely relies on loss trajectories rather than first- or second-order quantities (e.g., gradients and Hessians). 
To compare the impact measured by $\mathtt{LTC}$ to the influence measured by TDA metrics, we utilize the \textit{linear datamodeling score} ($\mathtt{LDS}$) introduced by \cite{trak_2023park}.
$\mathtt{LDS}$ measures the correlation between group-level attribution scores ($\mathtt{LTC}$ or influence) and their observed impact on model predictions when subsets of training data are used. 

\textbf{$\mathbf{LDS}$ definition} For a query data point $z_q$, random subsets $\{S_j\}_{j=1}^C$ are sampled from the training dataset, where each subset $S_j$ contains $\lceil \alpha N \rceil$ points, with $\alpha \in (0, 1)$ as the sampling ratio. 
Each subset $S_j$ is used to retrain the model $R$ times with different initializations $\{\xi_r\}_{r=1}^R$ and training parameters $\lambda$, resulting in the model $\theta^T_{S_j,\xi_r}$. 
This trained model is then used to compute a measurable quantity $f(\vec{z}_q, \theta^T_{S_j,\xi_r})$. 
A \textit{group attribution score}, $g_\tau(\vec{z}_q, S_j, S; \lambda)$, is calculated as $ g_\tau(\vec{z}_q, S_j, S; \lambda) \coloneqq \sum_{\vec{z} \in S_j} \tau(\vec{z}_q, \vec{z}, S; \lambda),$ where $\tau(\vec{z}_q, \vec{z}, S; \lambda)$ is the attribution score for a training point $\vec{z}$ with respect to $\vec{z}_q$. 
The $\mathtt{LDS}$ is then obtained using Spearman's rank~\cite{spearman1904} correlation ($\rho_s$):
{\small
\begin{align*}
\mathtt{LDS}(\vec{z}_q, \alpha) \coloneqq \rho_s \bigg( 
    & \left\{ \frac{1}{R} \sum_{r=1}^R f(\vec{z}_q, \theta^T_{S_j,\xi_r}) \right\}_{j=1}^C, \notag \\
    & \left\{ g_\tau(\vec{z}_q, S_j, S; \lambda) \right\}_{j=1}^C 
\bigg)
\end{align*}
}
\textbf{Experimental Setup:} We compared the $\mathtt{LDS}$ scores of $\mathtt{LTC}$ against those of $\mathtt{TRAK}$, $\mathtt{Arnoldi}$, $\mathtt{TracIn}$, $\mathtt{Infl}$, and $\mathtt{Datamodels}$. 
Precomputed scores for $\mathtt{Infl}$ and $\mathtt{Datamodels}$ were used for the CIFAR-10 dataset~\cite{cifar} with ResNet-9~\cite{resnet}, while 10 models were trained for $\mathtt{TRAK}$, $\mathtt{Arnoldi}$, $\mathtt{TracIn}$, and $\mathtt{LTC}$. 
The evaluation employed $C=100$ random subsets, sampling ratios $\alpha$ ranging from $0.3$ to $\frac{N-1}{N}$, a query set of 200 samples, and $R=10$ seeds. 
The measurable quantity was the accuracy of query samples.
\begin{figure}[t]
    \centering
    \includegraphics[width=1\linewidth]{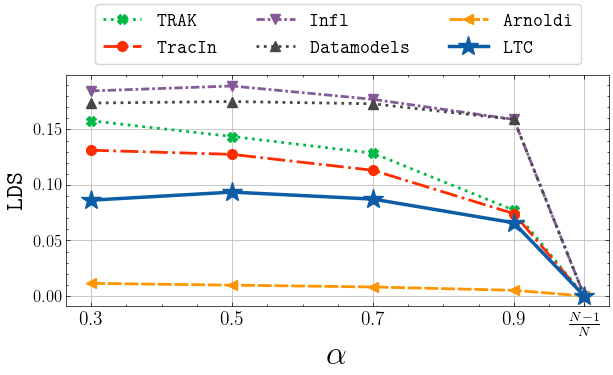}
    \vspace{-8mm}
    \caption{Linear datamodeling scores (LDS) of existing TDA metrics compared to $\mathtt{LTC}$. The scores were evaluated on CIFAR-10, ResNet-9 with 200 (randomly selected) query samples evaluated over 100 subsets.}
    \vspace{-4mm}
    \label{fig:lds_comparison}
\end{figure}

\textbf{Results and Observations:} The results presented in \cref{fig:lds_comparison} reveal that while the impact captured by $\mathtt{LTC}$ is distinct from the influence measured by traditional TDA metrics, it aligns closely with methods such as $\mathtt{TracIn}$ and $\mathtt{TRAK}$ in terms of behavior while being resource-efficient (explained in detail in \cref{appendix:TDA_metrics_overview}). 
Notably, the performance gap between these computationally intensive methods and $\mathtt{LTC}$ narrows as $\alpha$ increases.
The drop in $\mathtt{LDS}$ scores at $\alpha=\frac{N-1}{N}$ is due to the stochastic nature of model retraining\footnote{This observation is consistent with the findings of previous research \cite{revisitinglds_karthikeyan2021, source_bae2024}.}.

\textbf{Takeaways:} Although $\mathtt{LTC}$ fundamentally differs from influence-based TDA metrics, it mirrors their trends at higher sampling ratios while maintaining superior computational efficiency, solidifying its utility as a practical tool for analyzing training dynamics.

\subsection{Importance of the Top-$k$ Samples}
\label{sec:prediction_brittleness}
We demonstrate that the samples identified by $\mathtt{LTC}$ are indeed pivotal for generalization, addressing the question: \textit{“Are the training samples with the top-$k$ scores truly the most critical for forming a coreset?”} 
This is evaluated using the \textit{prediction brittleness} metric. 

\textbf{Experimental Setup:} To quantify the influence of top-$k$ samples, we systematically removed the most impactful data points identified by their $\mathtt{LTC}$ scores, from the training set and retrained the model. 
The metric of interest was the fraction of prediction flips observed in a held-out query set after retraining. 
If these samples are truly critical for generalization, their removal should cause substantial prediction changes. 
This experiment also included a comparative analysis with the top-$k$ influential samples identified by TDA scores, discussed in \cref{sec:TDA}. 
Experiments were performed on the CIFAR-10 dataset using a ResNet-9 architecture, with a randomly selected query set of 200 samples. 
For each configuration, once the top-$k$ samples were excluded, the model was retrained 5 times to account for randomness, and the average fraction of prediction flips was recorded.

\begin{figure}[t]
    \centering
    \includegraphics[width=1\linewidth]{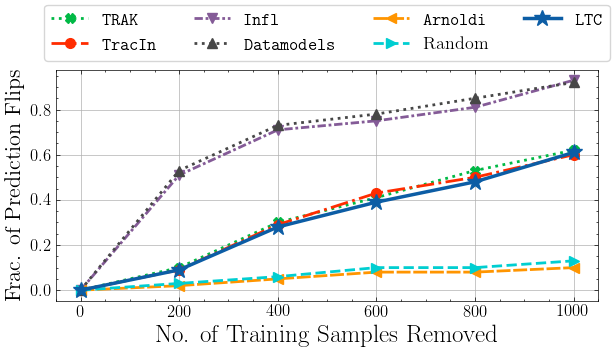}
    \vspace{-7mm}
    \caption{Prediction brittleness of $\mathtt{LTC}$ and TDA metrics on CIFAR-10 with ResNet-9. The top-$k$ influential training samples were removed, and the average prediction flips for 200 query samples over 5 seeds are shown.}
    \vspace{-6mm}
    \label{fig:prediction_brittleness}
\end{figure}

\textbf{Results and Observations:} The results, summarized in \cref{fig:prediction_brittleness}, illustrate that the top-$k$ samples identified by $\mathtt{LTC}$ have a comparable influence on prediction outcomes to those identified by TDA-based metrics such as $\mathtt{TracIn}$ and $\mathtt{TRAK}$. 
Notably, removing the top-800 samples of CIFAR-10, which constitutes just 1.6\% of the dataset, results in prediction flips for over half of the query set. 
This highlights the significant role of the samples identified by $\mathtt{LTC}$ in supporting model generalization. 
While metrics like $\mathtt{Datamodels}$ and $\mathtt{Infl}$ exhibit greater impact, they are computationally prohibitive, rendering them unsuitable for large-scale coreset generation. 

\textbf{Takeaways:} $\mathtt{LTC}$ emerges as an effective and computationally efficient approach for identifying training samples critical to generalization, making it a practical tool for coreset selection in large-scale machine learning pipelines.

\section{Limitations}
\label{sec:Limitations}
While $\mathtt{LTC}$ offers a scalable and effective method for quantifying the influence of training samples on test performance, it is not without limitations.
$\mathtt{LTC}$ assumes access to sufficient training checkpoints to monitor loss dynamics throughout training. 
This requirement can impose constraints in scenarios where training checkpoints are unavailable.

\section{Conclusion}
\label{sec:Conclusion}
In this work, we introduced Loss Trajectory Correlation ($\mathtt{LTC}$), a novel and efficient metric for coreset selection. 
By identifying training samples with high $\mathtt{LTC}$ values, our approach constructs compact and representative subsets that preserve generalization while reducing computational and memory demands. 
On CIFAR-100 and ImageNet-1k, LTC consistently delivers competitive accuracy, matching or outperforming state-of-the-art methods, and in the rare cases it doesn’t top the list, its performance is within $1\%$ of the best.
Moreover, $\mathtt{LTC}$-based coresets transfer effectively across diverse architectures, including ResNet, VGG, DenseNet, and Swin Transformer, with minimal performance degradation ($<2\%$). 
A significant advantage of $\mathtt{LTC}$ lies in its scalability. 
It achieves superior computational and storage efficiency compared to traditional methods, avoiding quadratic or high-dimensional dependencies, making it well-suited for large-scale datasets.
In addition to its efficiency, $\mathtt{LTC}$ provides insights into training dynamics, identifying aligned and conflicting samples at a low computational cost. 
This utility extends beyond coreset selection, offering a framework for dataset optimization and a better understanding of the role of training data in generalization.

\section*{Acknowledgements}
This work was supported in part by the Center for the Co-Design of Cognitive Systems (CoCoSys), a DARPA-sponsored JUMP 2.0 center, the Semiconductor Research Corporation (SRC), the National Science Foundation, and Collins Aerospace. We are also thankful to Amitangshu Mukherjee and Utkarsh Saxena for their helpful discussions and feedback.

\section*{Impact Statement}
\textbf
This paper presents a novel approach to coreset selection using Loss Trajectory Correlation ($\mathtt{LTC}$), which addresses critical challenges in efficient dataset optimization and scalable training. 
In the era of ever-growing datasets and complex models, our method offers a lightweight and scalable alternative to traditional training data attribution (TDA) techniques, requiring neither gradient computations nor expensive retraining. 
By leveraging naturally occurring loss trajectories, $\mathtt{LTC}$ facilitates the selection of compact yet representative coresets, ensuring computational efficiency without compromising model performance.

The broader implications of this work span multiple domains. $\mathtt{LTC}$ promotes sustainability by reducing the computational and memory overheads associated with large-scale machine learning workflows, thereby lowering energy consumption and environmental impact. 
Additionally, by identifying the most impactful training samples, $\mathtt{LTC}$ can reveal hidden biases within datasets, providing opportunities to mitigate these biases and improve fairness in model training. 
The proposed approach tackles a realistic and essential problem in deep learning, requiring advanced techniques and a nuanced understanding of training dynamics. 
While introducing a significant step forward in coreset generation, this work adheres to ethical standards and avoids negative implications, ensuring its contributions are both impactful and responsible.

\bibliography{main}
\bibliographystyle{icml2025}
\appendix
\section{Detailed Results for Coreset Generation Experiments}
\label{appendix:coreset_gen}
\paragraph{CIFAR-100} All networks were trained using an SGD optimizer \cite{sgd} for $164$ epochs with a learning rate of $0.1$, scaled by $0.1$ at epochs $81$ and $121$. 
Nesterov momentum \citep{nesterov}, with momentum of $0.9$, was turned on and a weight decay of $5\text{e}-4$ was used. 
We used the following sequence of data augmentations for training: resize to ($32 \times 32$), random crop with padding $=4$, random horizontal flip, and normalization.

\paragraph{ImageNet-1k} The training setup for ImageNet-1k mirrored best practices for large-scale datasets. All networks were trained using an SGD optimizer for 90 epochs, with an initial learning rate of $0.1$, reduced by a factor of $0.1$ at epochs 30 and 60. Nesterov momentum, with a momentum coefficient of $0.9$, was applied, alongside a weight decay of $1\text{e}-4$. The training employed data augmentations including random resized cropping ($224 \times 224$), random horizontal flipping, and normalization. 

No additional finetuning or regularizing was utilized for any of the coreset generation techniques (including ours). 
This was done to ensure complete fairness in comparison. 
All methods used the entire test set as the validation set to choose samples for the coreset.

Each experiment was conducted over 5 independent runs with distinct random seeds, and the mean and variance of results are summarized in Tables \ref{tab:coresets_cifar100} and \ref{tab:coresets_imagenet}. 
\begin{table*}[ht]
\centering
\caption{Performance summary of various coreset generation techniques on CIFAR-100 with ResNet-18. The mean values are shown in the first line and the standard deviations are shown in the second. The highest values for each coreset sizes are highlighted in bold. }
\label{tab:coresets_cifar100}
\vspace{2mm}
\renewcommand{\arraystretch}{1.1}
\begin{tabular}{c|ccccccccc}
\hline
\begin{tabular}[c]{@{}c@{}}Coreset size\\ (\%. dataset)\end{tabular} &
Random &
$\mathtt{Cal}$ &
$\mathtt{GraphCut}$ &
$\mathtt{Glister}$ &
$\mathtt{GraNd}$ &
$\mathtt{Forgetting}$ &
$\mathtt{Herding}$ &
$\mathtt{Slocurv}$ &
\cellcolor[HTML]{E6E6FF} $\mathtt{LTC}$\\ \hline \hline

0.2\% & \begin{tabular}[c]{@{}c@{}}3.66 \\ $\pm 0.41$\end{tabular} & \begin{tabular}[c]{@{}c@{}}5.24 \\ $\pm 0.41$\end{tabular} & \begin{tabular}[c]{@{}c@{}} \textbf{5.80} \\ $\mathbf{\pm 0.24}$\end{tabular} & \begin{tabular}[c]{@{}c@{}}3.43 \\ $\pm 0.32$\end{tabular} & \begin{tabular}[c]{@{}c@{}}2.49 \\ $\pm 0.20$\end{tabular} & \begin{tabular}[c]{@{}c@{}}3.52 \\ $\pm 0.16$\end{tabular} & \begin{tabular}[c]{@{}c@{}}2.57 \\ $\pm 0.52$\end{tabular} & \begin{tabular}[c]{@{}c@{}}3.62 \\ $\pm 0.44$\end{tabular} &\cellcolor[HTML]{E6E6FF} \begin{tabular}[c]{@{}c@{}}5.18 \\ $\pm 0.36$\end{tabular} \\ \hline

0.4\% & \begin{tabular}[c]{@{}c@{}}6.03 \\ $\pm 0.28$\end{tabular} & \begin{tabular}[c]{@{}c@{}}\textbf{7.46} \\ $\mathbf{\pm 0.28}$\end{tabular} & \begin{tabular}[c]{@{}c@{}}7.07 \\ $\pm 0.39$\end{tabular} & \begin{tabular}[c]{@{}c@{}}4.91 \\ $\pm 0.37$\end{tabular} & \begin{tabular}[c]{@{}c@{}}3.67 \\ $\pm 0.29$\end{tabular} & \begin{tabular}[c]{@{}c@{}}5.12 \\ $\pm 0.53$\end{tabular} & \begin{tabular}[c]{@{}c@{}}3.42 \\ $\pm 0.49$\end{tabular} & \begin{tabular}[c]{@{}c@{}}5.46 \\ $\pm 0.36$\end{tabular} &\cellcolor[HTML]{E6E6FF} \begin{tabular}[c]{@{}c@{}}7.43 \\ $\pm 0.43$\end{tabular} \\ \hline

0.6\% & \begin{tabular}[c]{@{}c@{}}6.80 \\ $\pm 0.27$\end{tabular} & \begin{tabular}[c]{@{}c@{}}\textbf{9.12} \\ $\mathbf{\pm 0.27}$\end{tabular} & \begin{tabular}[c]{@{}c@{}}8.96 \\ $\pm 0.31$\end{tabular} & \begin{tabular}[c]{@{}c@{}}7.49 \\ $\pm 0.61$\end{tabular} & \begin{tabular}[c]{@{}c@{}}5.61 \\ $\pm 0.14$\end{tabular} & \begin{tabular}[c]{@{}c@{}}6.80 \\ $\pm 0.18$\end{tabular} & \begin{tabular}[c]{@{}c@{}}4.07 \\ $\pm 0.41$\end{tabular} & \begin{tabular}[c]{@{}c@{}}7.44 \\ $\pm 0.20$\end{tabular} &\cellcolor[HTML]{E6E6FF} \begin{tabular}[c]{@{}c@{}}8.76 \\ $\pm 0.40$\end{tabular} \\ \hline

0.8\% & \begin{tabular}[c]{@{}c@{}}7.96 \\ $\pm 0.79$\end{tabular} & \begin{tabular}[c]{@{}c@{}}\textbf{10.19} \\ $\mathbf{\pm 0.79}$\end{tabular} & \begin{tabular}[c]{@{}c@{}}9.39 \\ $\pm 0.81$\end{tabular} & \begin{tabular}[c]{@{}c@{}}8.65 \\ $\pm 0.47$\end{tabular} & \begin{tabular}[c]{@{}c@{}}7.03 \\ $\pm 0.52$\end{tabular} & \begin{tabular}[c]{@{}c@{}}8.42 \\ $\pm 0.38$\end{tabular} & \begin{tabular}[c]{@{}c@{}}5.14 \\ $\pm 0.70$\end{tabular} & \begin{tabular}[c]{@{}c@{}}9.96 \\ $\pm 0.42$\end{tabular} & \cellcolor[HTML]{E6E6FF} \begin{tabular}[c]{@{}c@{}}9.88 \\ $\pm 0.48$\end{tabular} \\ \hline

1\% & \begin{tabular}[c]{@{}c@{}}9.38 \\ $\pm 0.41$\end{tabular} & \begin{tabular}[c]{@{}c@{}}\textbf{13.73} \\ $\mathbf{\pm 0.41}$\end{tabular} & \begin{tabular}[c]{@{}c@{}}11.85 \\ $\pm 0.47$\end{tabular} & \begin{tabular}[c]{@{}c@{}}9.04 \\ $\pm 0.43$\end{tabular} & \begin{tabular}[c]{@{}c@{}}9.68 \\ $\pm 0.17$\end{tabular} & \begin{tabular}[c]{@{}c@{}}11.53 \\ $\pm 0.43$\end{tabular} & \begin{tabular}[c]{@{}c@{}}5.31 \\ $\pm 0.32$\end{tabular} & \begin{tabular}[c]{@{}c@{}}12.17 \\ $\pm 0.30$\end{tabular} & \cellcolor[HTML]{E6E6FF} \begin{tabular}[c]{@{}c@{}}11.56 \\ $\pm 0.36$\end{tabular} \\ \hline

2\% & \begin{tabular}[c]{@{}c@{}}12.74 \\ $\pm 0.28$\end{tabular} & \begin{tabular}[c]{@{}c@{}}16.45 \\ $\pm 0.28$\end{tabular} & \begin{tabular}[c]{@{}c@{}}16.95 \\ $\pm 0.55$\end{tabular} & \begin{tabular}[c]{@{}c@{}}14.54 \\ $\pm 0.41$\end{tabular} & \begin{tabular}[c]{@{}c@{}}14.63 \\ $\pm 0.17$\end{tabular} & \begin{tabular}[c]{@{}c@{}}15.89 \\ $\pm 0.21$\end{tabular} & \begin{tabular}[c]{@{}c@{}}8.29 \\ $\pm 0.45$\end{tabular} & \begin{tabular}[c]{@{}c@{}}13.34 \\ $\pm 0.42$\end{tabular} & \cellcolor[HTML]{E6E6FF} \begin{tabular}[c]{@{}c@{}}\textbf{16.77} \\ $\mathbf{\pm 0.56}$\end{tabular} \\ \hline

3\% & \begin{tabular}[c]{@{}c@{}}16.58 \\ $\pm 0.95$\end{tabular} & \begin{tabular}[c]{@{}c@{}}20.05 \\ $\pm 0.95$\end{tabular} & \begin{tabular}[c]{@{}c@{}}19.21 \\ $\pm 0.57$\end{tabular} & \begin{tabular}[c]{@{}c@{}}17.47 \\ $\pm 0.32$\end{tabular} & \begin{tabular}[c]{@{}c@{}}16.71 \\ $\pm 0.14$\end{tabular} & \begin{tabular}[c]{@{}c@{}}18.24 \\ $\pm 0.32$\end{tabular} & \begin{tabular}[c]{@{}c@{}}9.23 \\ $\pm 0.52$\end{tabular} & \begin{tabular}[c]{@{}c@{}}\textbf{22.67} \\ $\mathbf{\pm 0.80}$\end{tabular} & \cellcolor[HTML]{E6E6FF} \begin{tabular}[c]{@{}c@{}}19.75 \\ $\pm 0.87$\end{tabular} \\ \hline

4\% & \begin{tabular}[c]{@{}c@{}}19.99 \\ $\pm 1.03$\end{tabular} & \begin{tabular}[c]{@{}c@{}}22.97 \\ $\pm 1.03$\end{tabular} & \begin{tabular}[c]{@{}c@{}}21.33 \\ $\pm 0.71$\end{tabular} & \begin{tabular}[c]{@{}c@{}}\textbf{23.99} \\ $\mathbf{\pm 0.58}$\end{tabular} & \begin{tabular}[c]{@{}c@{}}22.33 \\ $\pm 0.31$\end{tabular} & \begin{tabular}[c]{@{}c@{}}23.82 \\ $\pm 0.86$\end{tabular} & \begin{tabular}[c]{@{}c@{}}11.52 \\ $\pm 1.15$\end{tabular} & \begin{tabular}[c]{@{}c@{}}21.97 \\ $\pm 0.55$\end{tabular} & \cellcolor[HTML]{E6E6FF} \begin{tabular}[c]{@{}c@{}}22.32 \\ $\pm 0.82$\end{tabular} \\ \hline

5\% & \begin{tabular}[c]{@{}c@{}}22.41 \\ $\pm 0.54$\end{tabular} & \begin{tabular}[c]{@{}c@{}}24.37 \\ $\pm 0.54$\end{tabular} & \begin{tabular}[c]{@{}c@{}}26.31 \\ $\pm 0.58$\end{tabular} & \begin{tabular}[c]{@{}c@{}}24.82 \\ $\pm 0.72$\end{tabular} & \begin{tabular}[c]{@{}c@{}}22.10 \\ $\pm 0.48$\end{tabular} & \begin{tabular}[c]{@{}c@{}}26.38 \\ $\pm 0.84$\end{tabular} & \begin{tabular}[c]{@{}c@{}}13.66 \\ $\pm 1.35$\end{tabular} & \begin{tabular}[c]{@{}c@{}}23.44 \\ $\pm 0.71$\end{tabular} & \cellcolor[HTML]{E6E6FF} \begin{tabular}[c]{@{}c@{}}\textbf{26.33} \\ $\mathbf{\pm 1.40}$\end{tabular} \\ \hline

6\% & \begin{tabular}[c]{@{}c@{}}23.66 \\ $\pm 0.49$\end{tabular} & \begin{tabular}[c]{@{}c@{}}26.93 \\ $\pm 0.49$\end{tabular} & \begin{tabular}[c]{@{}c@{}}\textbf{30.35} \\ $\mathbf{\pm 0.68}$\end{tabular} & \begin{tabular}[c]{@{}c@{}}26.57 \\ $\pm 0.69$\end{tabular} & \begin{tabular}[c]{@{}c@{}}23.01 \\ $\pm 0.47$\end{tabular} & \begin{tabular}[c]{@{}c@{}}28.16 \\ $\pm 0.62$\end{tabular} & \begin{tabular}[c]{@{}c@{}}15.49 \\ $\pm 0.91$\end{tabular} & \begin{tabular}[c]{@{}c@{}}25.41 \\ $\pm 0.43$\end{tabular} & \cellcolor[HTML]{E6E6FF} \begin{tabular}[c]{@{}c@{}}30.19 \\ $\pm 0.92$\end{tabular} \\ \hline

7\% & \begin{tabular}[c]{@{}c@{}}28.36 \\ $\pm 0.85$\end{tabular} & \begin{tabular}[c]{@{}c@{}}27.37 \\ $\pm 0.85$\end{tabular} & \begin{tabular}[c]{@{}c@{}}\textbf{31.63} \\ $\mathbf{\pm 0.71}$\end{tabular} & \begin{tabular}[c]{@{}c@{}}27.57 \\ $\pm 0.84$\end{tabular} & \begin{tabular}[c]{@{}c@{}}25.40 \\ $\pm 0.24$\end{tabular} & \begin{tabular}[c]{@{}c@{}}30.95 \\ $\pm 0.66$\end{tabular} & \begin{tabular}[c]{@{}c@{}}18.52 \\ $\pm 0.56$\end{tabular} & \begin{tabular}[c]{@{}c@{}}27.45 \\ $\pm 0.60$\end{tabular} & \cellcolor[HTML]{E6E6FF} \begin{tabular}[c]{@{}c@{}}28.82 \\ $\pm 0.92$\end{tabular} \\ \hline

8\% & \begin{tabular}[c]{@{}c@{}}30.75 \\ $\pm 1.12$\end{tabular} & \begin{tabular}[c]{@{}c@{}}28.32 \\ $\pm 1.12$\end{tabular} & \begin{tabular}[c]{@{}c@{}}32.22 \\ $\pm 0.48$\end{tabular} & \begin{tabular}[c]{@{}c@{}}28.79 \\ $\pm 0.79$\end{tabular} & \begin{tabular}[c]{@{}c@{}}26.47 \\ $\pm 0.47$\end{tabular} & \begin{tabular}[c]{@{}c@{}}31.84 \\ $\pm 0.23$\end{tabular} & \begin{tabular}[c]{@{}c@{}}18.52 \\ $\pm 0.39$\end{tabular} & \begin{tabular}[c]{@{}c@{}}29.17 \\ $\pm 0.47$\end{tabular} & \cellcolor[HTML]{E6E6FF} \begin{tabular}[c]{@{}c@{}}\textbf{32.40} \\ $\mathbf{\pm 0.90}$\end{tabular} \\ \hline

9\% & \begin{tabular}[c]{@{}c@{}}32.12 \\ $\pm 1.48$\end{tabular} & \begin{tabular}[c]{@{}c@{}}29.19 \\ $\pm 1.48$\end{tabular} & \begin{tabular}[c]{@{}c@{}}33.01 \\ $\pm 0.82$\end{tabular} & \begin{tabular}[c]{@{}c@{}}30.21 \\ $\pm 0.32$\end{tabular} & \begin{tabular}[c]{@{}c@{}}27.66 \\ $\pm 0.39$\end{tabular} & \begin{tabular}[c]{@{}c@{}}32.79 \\ $\pm 0.94$\end{tabular} & \begin{tabular}[c]{@{}c@{}}18.52 \\ $\pm 0.98$\end{tabular} & \begin{tabular}[c]{@{}c@{}}30.71 \\ $\pm 0.57$\end{tabular} & \cellcolor[HTML]{E6E6FF} \begin{tabular}[c]{@{}c@{}}\textbf{33.89} \\ $\mathbf{\pm 0.80}$\end{tabular} \\ \hline

10\% & \begin{tabular}[c]{@{}c@{}}32.75 \\ $\pm 1.02$\end{tabular} & \begin{tabular}[c]{@{}c@{}}31.02 \\ $\pm 1.02$\end{tabular} & \begin{tabular}[c]{@{}c@{}}34.41 \\ $\pm 0.96$\end{tabular} & \begin{tabular}[c]{@{}c@{}}31.22 \\ $\pm 1.33$\end{tabular} & \begin{tabular}[c]{@{}c@{}}29.45 \\ $\pm 0.51$\end{tabular} & \begin{tabular}[c]{@{}c@{}}33.04 \\ $\pm 0.65$\end{tabular} & \begin{tabular}[c]{@{}c@{}}19.54 \\ $\pm 0.85$\end{tabular} & \begin{tabular}[c]{@{}c@{}}33.17 \\ $\pm 1.16$\end{tabular} & \cellcolor[HTML]{E6E6FF} \begin{tabular}[c]{@{}c@{}}\textbf{35.47} \\ $\mathbf{\pm 0.35}$\end{tabular} \\ \hline

\end{tabular}
\end{table*}

\begin{table*}[ht]
\centering
\caption{Performance summary of various coreset generation techniques on ImageNet-1k with ResNet-18. The mean values are shown in the first line and the standard deviations are shown in the second. The highest values for each coreset sizes are highlighted in bold.}
\label{tab:coresets_imagenet}
\vspace{2mm}
\renewcommand{\arraystretch}{1.1}
\begin{tabular}{c|ccccccccc}
\hline
\begin{tabular}[c]{@{}c@{}}Coreset size\\ (\%. dataset)\end{tabular} &
Random &
$\mathtt{Cal}$ &
$\mathtt{GraphCut}$ &
$\mathtt{Glister}$ &
$\mathtt{GraNd}$ &
$\mathtt{Forgetting}$ &
$\mathtt{Herding}$ &
$\mathtt{Slocurv}$ &
\cellcolor[HTML]{E6E6FF} \begin{tabular}[c]{@{}c@{}}Ours \\ $\mathtt{LTC}$\end{tabular} \\ \hline \hline

0.1\% & \begin{tabular}[c]{@{}c@{}}0.70 \\ $\pm 0.03$\end{tabular} & \begin{tabular}[c]{@{}c@{}}1.13 \\ $\pm 0.12$\end{tabular} & \begin{tabular}[c]{@{}c@{}}1.09 \\ $\pm 0.09$\end{tabular} & \begin{tabular}[c]{@{}c@{}}0.86 \\ $\pm 0.01$\end{tabular} & \begin{tabular}[c]{@{}c@{}}0.97 \\ $\pm 0.01$\end{tabular} & \begin{tabular}[c]{@{}c@{}}0.64 \\ $\pm 0.01$\end{tabular} & \begin{tabular}[c]{@{}c@{}}0.31 \\ $\pm 0.01$\end{tabular} & \begin{tabular}[c]{@{}c@{}}1.23 \\ $\pm 0.06$\end{tabular} & \cellcolor[HTML]{E6E6FF} \begin{tabular}[c]{@{}c@{}}\textbf{1.95} \\ $\mathbf{\pm 0.90}$\end{tabular} \\ \hline

0.5\% & \begin{tabular}[c]{@{}c@{}}3.98 \\ $\pm 0.19$\end{tabular} & \begin{tabular}[c]{@{}c@{}}6.84 \\ $\pm 0.13$\end{tabular} & \begin{tabular}[c]{@{}c@{}}\textbf{7.27} \\ $\mathbf{\pm 0.03}$\end{tabular} & \begin{tabular}[c]{@{}c@{}}5.55 \\ $\pm 0.05$\end{tabular} & \begin{tabular}[c]{@{}c@{}}6.81 \\ $\pm 0.20$\end{tabular} & \begin{tabular}[c]{@{}c@{}}4.78 \\ $\pm 1.01$\end{tabular} & \begin{tabular}[c]{@{}c@{}}1.39 \\ $\pm 0.17$\end{tabular} & \begin{tabular}[c]{@{}c@{}}5.90 \\ $\pm 0.07$\end{tabular} & \cellcolor[HTML]{E6E6FF} \begin{tabular}[c]{@{}c@{}}6.91 \\ $\pm 0.56$\end{tabular} \\ \hline

1\% & \begin{tabular}[c]{@{}c@{}}7.86 \\ $\pm 0.43$\end{tabular} & \begin{tabular}[c]{@{}c@{}}13.17 \\ $\pm 0.22$\end{tabular} & \begin{tabular}[c]{@{}c@{}}14.27 \\ $\pm 0.31$\end{tabular} & \begin{tabular}[c]{@{}c@{}}12.45 \\ $\pm 0.01$\end{tabular} & \begin{tabular}[c]{@{}c@{}}16.07 \\ $\pm 0.23$\end{tabular} & \begin{tabular}[c]{@{}c@{}}12.67 \\ $\pm 0.51$\end{tabular} & \begin{tabular}[c]{@{}c@{}}4.32 \\ $\pm 0.62$\end{tabular} & \begin{tabular}[c]{@{}c@{}}14.17 \\ $\pm 0.02$\end{tabular} & \cellcolor[HTML]{E6E6FF} \begin{tabular}[c]{@{}c@{}}\textbf{15.89} \\ $\mathbf{\pm 1.21}$\end{tabular} \\ \hline

5\% & \begin{tabular}[c]{@{}c@{}}39.78 \\ $\pm 0.23$\end{tabular} & \begin{tabular}[c]{@{}c@{}}37.65 \\ $\pm 1.30$\end{tabular} & \begin{tabular}[c]{@{}c@{}}39.80 \\ $\pm 0.60$\end{tabular} & \begin{tabular}[c]{@{}c@{}}42.19 \\ $\pm 0.03$\end{tabular} & \begin{tabular}[c]{@{}c@{}}41.41 \\ $\pm 0.14$\end{tabular} & \begin{tabular}[c]{@{}c@{}}44.85 \\ $\pm 0.74$\end{tabular} & \begin{tabular}[c]{@{}c@{}}15.36 \\ $\pm 0.18$\end{tabular} & \begin{tabular}[c]{@{}c@{}}40.10 \\ $\pm 0.14$\end{tabular} & \cellcolor[HTML]{E6E6FF} \begin{tabular}[c]{@{}c@{}}\textbf{45.15} \\ $\mathbf{\pm 0.01}$\end{tabular} \\ \hline

10\% & \begin{tabular}[c]{@{}c@{}}51.24 \\ $\pm 0.04$\end{tabular} & \begin{tabular}[c]{@{}c@{}}44.16 \\ $\pm 0.78$\end{tabular} & \begin{tabular}[c]{@{}c@{}}48.27 \\ $\pm 1.02$\end{tabular} & \begin{tabular}[c]{@{}c@{}}50.10 \\ $\pm 0.01$\end{tabular} & \begin{tabular}[c]{@{}c@{}}47.28 \\ $\pm 0.56$\end{tabular} & \begin{tabular}[c]{@{}c@{}}53.19 \\ $\pm 0.06$\end{tabular} & \begin{tabular}[c]{@{}c@{}}26.84 \\ $\pm 0.05$\end{tabular} & \begin{tabular}[c]{@{}c@{}}46.39 \\ $\pm 0.50$\end{tabular} & \cellcolor[HTML]{E6E6FF} \begin{tabular}[c]{@{}c@{}}\textbf{53.78} \\ $\mathbf{\pm 0.30}$\end{tabular} \\ \hline

30\% & \begin{tabular}[c]{@{}c@{}}60.87 \\ $\pm 0.13$\end{tabular} & \begin{tabular}[c]{@{}c@{}}54.41 \\ $\pm 0.45$\end{tabular} & \begin{tabular}[c]{@{}c@{}}\textbf{61.24} \\ $\mathbf{\pm 0.01}$\end{tabular} & \begin{tabular}[c]{@{}c@{}}58.52 \\ $\pm 0.05$\end{tabular} & \begin{tabular}[c]{@{}c@{}}55.76 \\ $\pm 0.70$\end{tabular} & \begin{tabular}[c]{@{}c@{}}60.90 \\ $\pm 0.05$\end{tabular} & \begin{tabular}[c]{@{}c@{}}46.61 \\ $\pm 0.87$\end{tabular} & \begin{tabular}[c]{@{}c@{}}57.19 \\ $\pm 0.01$\end{tabular} & \cellcolor[HTML]{E6E6FF} \begin{tabular}[c]{@{}c@{}}61.11 \\ $\pm 0.41$\end{tabular} \\ \hline
\end{tabular}
\end{table*}

\begin{table*}[ht]
\centering
\caption{Cross-architecture performance of coresets of different sizes of ImageNet-1k identified by $\mathtt{LTC}$. The mean values are shown in the first line of each cell and the standard deviations are shown in the second. The results demonstrate the transferability of coresets identified using a smaller source model (ResNet-18) as there is a minimal accuracy drop ($<2\%$) compared to coresets identified using source models having the same architecture as the target model.}
\vspace{1mm}
\label{tab:transferability}
\renewcommand{\arraystretch}{1.25}
\begin{tabular}{cc|cccccc}
\cline{3-8}
\multicolumn{2}{c|}{} &
  \multicolumn{6}{c}{Coreset size (\%. of dataset)} \\ 
\multicolumn{1}{c}{Target Model} &
  Source Model &
  0.1\% &
  0.5\% &
  1\% &
  5\% &
  10\% &
  30\% \\ \hline \hline
\multicolumn{1}{c|}{} &
  \cellcolor[HTML]{E6E6FF}ResNet-18 &
  \cellcolor[HTML]{E6E6FF}\begin{tabular}[c]{@{}c@{}}1.81\\  $\pm$  0.05\end{tabular} &
  \cellcolor[HTML]{E6E6FF}\begin{tabular}[c]{@{}c@{}}6.91\\  $\pm$  0.51\end{tabular} &
  \cellcolor[HTML]{E6E6FF}\begin{tabular}[c]{@{}c@{}}16.98\\  $\pm$  0.43\end{tabular} &
  \cellcolor[HTML]{E6E6FF}\begin{tabular}[c]{@{}c@{}}46.21\\  $\pm$  0.57\end{tabular} &
  \cellcolor[HTML]{E6E6FF}\begin{tabular}[c]{@{}c@{}}54.13\\  $\pm$  0.03\end{tabular} &
  \cellcolor[HTML]{E6E6FF}\begin{tabular}[c]{@{}c@{}}63.81\\  $\pm$  0.91\end{tabular} \\
\multicolumn{1}{c|}{\multirow{-2}{*}{ResNet-34}} &
  ResNet-34 &
  \begin{tabular}[c]{@{}c@{}}1.90\\  $\pm$  0.10\end{tabular} &
  \begin{tabular}[c]{@{}c@{}}7.03\\  $\pm$  0.28\end{tabular} &
  \begin{tabular}[c]{@{}c@{}}17.56\\  $\pm$  0.10\end{tabular} &
  \begin{tabular}[c]{@{}c@{}}47.08\\  $\pm$  0.41\end{tabular} &
  \begin{tabular}[c]{@{}c@{}}55.31\\  $\pm$  0.14\end{tabular} &
  \begin{tabular}[c]{@{}c@{}}63.78\\  $\pm$  0.02\end{tabular} \\ \hline \hline
\multicolumn{1}{c|}{} &
  \cellcolor[HTML]{E6E6FF}ResNet-18 &
  \cellcolor[HTML]{E6E6FF}\begin{tabular}[c]{@{}c@{}}4.12\\  $\pm$  0.01\end{tabular} &
  \cellcolor[HTML]{E6E6FF}\begin{tabular}[c]{@{}c@{}}9.31\\  $\pm$  0.45\end{tabular} &
  \cellcolor[HTML]{E6E6FF}\begin{tabular}[c]{@{}c@{}}18.92\\  $\pm$  1.23\end{tabular} &
  \cellcolor[HTML]{E6E6FF}\begin{tabular}[c]{@{}c@{}}50.04\\  $\pm$  0.01\end{tabular} &
  \cellcolor[HTML]{E6E6FF}\begin{tabular}[c]{@{}c@{}}57.91\\  $\pm$  0.13\end{tabular} &
  \cellcolor[HTML]{E6E6FF}\begin{tabular}[c]{@{}c@{}}67.02\\  $\pm$  0.87\end{tabular} \\
\multicolumn{1}{c|}{\multirow{-2}{*}{ResNet-50}} &
  ResNet-50 &
  \begin{tabular}[c]{@{}c@{}}4.38\\  $\pm$  0.50\end{tabular} &
  \begin{tabular}[c]{@{}c@{}}9.67\\  $\pm$  0.13\end{tabular} &
  \begin{tabular}[c]{@{}c@{}}19.14\\  $\pm$  0.10\end{tabular} &
  \begin{tabular}[c]{@{}c@{}}50.14\\  $\pm$  0.27\end{tabular} &
  \begin{tabular}[c]{@{}c@{}}58.64\\  $\pm$  0.14\end{tabular} &
  \begin{tabular}[c]{@{}c@{}}66.37\\  $\pm$  0.01\end{tabular} \\ \hline \hline
\multicolumn{1}{c|}{} &
  \cellcolor[HTML]{E6E6FF}ResNet-18 &
  \cellcolor[HTML]{E6E6FF}\begin{tabular}[c]{@{}c@{}}3.89\\  $\pm$  0.12\end{tabular} &
  \cellcolor[HTML]{E6E6FF}\begin{tabular}[c]{@{}c@{}}13.02\\  $\pm$  0.06\end{tabular} &
  \cellcolor[HTML]{E6E6FF}\begin{tabular}[c]{@{}c@{}}24.50\\  $\pm$  0.51\end{tabular} &
  \cellcolor[HTML]{E6E6FF}\begin{tabular}[c]{@{}c@{}}55.92\\  $\pm$  0.78\end{tabular} &
  \cellcolor[HTML]{E6E6FF}\begin{tabular}[c]{@{}c@{}}62.12\\  $\pm$  1.23\end{tabular} &
  \cellcolor[HTML]{E6E6FF}\begin{tabular}[c]{@{}c@{}}69.97\\  $\pm$  0.40\end{tabular} \\
\multicolumn{1}{c|}{\multirow{-2}{*}{DenseNet-121}} &
  DenseNet-121 &
  \begin{tabular}[c]{@{}c@{}}4.21\\  $\pm$  0.17\end{tabular} &
  \begin{tabular}[c]{@{}c@{}}15.02\\  $\pm$  0.31\end{tabular} &
  \begin{tabular}[c]{@{}c@{}}25.07\\  $\pm$  0.02\end{tabular} &
  \begin{tabular}[c]{@{}c@{}}57.03\\  $\pm$  0.10\end{tabular} &
  \begin{tabular}[c]{@{}c@{}}64.21\\  $\pm$  0.05\end{tabular} &
  \begin{tabular}[c]{@{}c@{}}71.19\\  $\pm$  1.03\end{tabular} \\ \hline \hline
\multicolumn{1}{c|}{} &
  \cellcolor[HTML]{E6E6FF}ResNet-18 &
  \cellcolor[HTML]{E6E6FF}\begin{tabular}[c]{@{}c@{}}3.76\\  $\pm$  0.14\end{tabular} &
  \cellcolor[HTML]{E6E6FF}\begin{tabular}[c]{@{}c@{}}14.41\\  $\pm$  0.23\end{tabular} &
  \cellcolor[HTML]{E6E6FF}\begin{tabular}[c]{@{}c@{}}22.50\\  $\pm$  0.12\end{tabular} &
  \cellcolor[HTML]{E6E6FF}\begin{tabular}[c]{@{}c@{}}54.03\\  $\pm$  0.67\end{tabular} &
  \cellcolor[HTML]{E6E6FF}\begin{tabular}[c]{@{}c@{}}63.12\\  $\pm$  1.05\end{tabular} &
  \cellcolor[HTML]{E6E6FF}\begin{tabular}[c]{@{}c@{}}71.68\\  $\pm$  0.84\end{tabular} \\
\multicolumn{1}{c|}{\multirow{-2}{*}{VGG-19(bn)}} &
  VGG-19(bn) &
  \begin{tabular}[c]{@{}c@{}}3.81\\  $\pm$  0.14\end{tabular} &
  \begin{tabular}[c]{@{}c@{}}15.68\\  $\pm$  0.04\end{tabular} &
  \begin{tabular}[c]{@{}c@{}}23.71\\  $\pm$  0.13\end{tabular} &
  \begin{tabular}[c]{@{}c@{}}55.89\\  $\pm$  0.56\end{tabular} &
  \begin{tabular}[c]{@{}c@{}}62.68\\  $\pm$  0.43\end{tabular} &
  \begin{tabular}[c]{@{}c@{}}72.01\\  $\pm$  0.52\end{tabular} \\ \hline \hline
\multicolumn{1}{c|}{} &
  \cellcolor[HTML]{E6E6FF}ResNet-18 &
  \cellcolor[HTML]{E6E6FF}\begin{tabular}[c]{@{}c@{}}3.01\\  $\pm$  0.02\end{tabular} &
  \cellcolor[HTML]{E6E6FF}\begin{tabular}[c]{@{}c@{}}17.75\\  $\pm$  0.31\end{tabular} &
  \cellcolor[HTML]{E6E6FF}\begin{tabular}[c]{@{}c@{}}26.85\\  $\pm$  0.06\end{tabular} &
  \cellcolor[HTML]{E6E6FF}\begin{tabular}[c]{@{}c@{}}58.29\\  $\pm$  0.51\end{tabular} &
  \cellcolor[HTML]{E6E6FF}\begin{tabular}[c]{@{}c@{}}62.18\\  $\pm$  0.17\end{tabular} &
  \cellcolor[HTML]{E6E6FF}\begin{tabular}[c]{@{}c@{}}73.12\\  $\pm$  0.65\end{tabular} \\
\multicolumn{1}{c|}{\multirow{-2}{*}{Swin-T}} &
  Swin-T &
  \begin{tabular}[c]{@{}c@{}}2.12\\  $\pm$  0.92\end{tabular} &
  \begin{tabular}[c]{@{}c@{}}18.21\\  $\pm$  0.31\end{tabular} &
  \begin{tabular}[c]{@{}c@{}}27.61\\  $\pm$  0.76\end{tabular} &
  \begin{tabular}[c]{@{}c@{}}59.81\\  $\pm$  0.53\end{tabular} &
  \begin{tabular}[c]{@{}c@{}}64.96\\  $\pm$  0.83\end{tabular} &
  \begin{tabular}[c]{@{}c@{}}75.43\\  $\pm$  0.59\end{tabular} \\ \hline
\end{tabular}
\end{table*}

The results of the cross-architecture performance that demonstrates the transferability of the coresets generated using $\mathtt{LTC}$, discussed in \cref{sec:transferability} is shown in \cref{tab:transferability}.

\section{Detailed Explanation of Overheads from Coreset Methodologies}
\label{appendix:details_overheads_coresets}
\paragraph{Recap on Notation} We denote the number of training samples as $N$ ($S \sim \mathbf{Z}^N$) and the number of query (test/validation) samples as $Q$. 
The model is trained for $T$ epochs, and for certain TDA metrics, there are hyperparameters (denoted as $\lambda_\tau$) that are used in calculating the TDA metric $\tau$ that further influence the computational cost. 
The computational complexity of each method is expressed in terms of floating point operations per second (FLOPs), denoted as $f$, which represent the cost of a single forward pass for a model with $p$ parameters. 
The cost of a backward pass is approximately $2f$ FLOPs.
The storage overhead is expressed in terms of the number of parameters $p$ of the model. 
$R$ is the number of model retrainings required. 
$k$ is the size of the coreset that needs to be generated, while $\gamma$ is the frequency of coreset generation. $d$ is the feature dimensionality of samples from the model. 

\paragraph{Generalization-based Data Subset Selection for Efficient and Robust Learning ($\mathtt{Glister}$~\cite{glister_killamsetty2021}} formulates coreset selection as a mixed discrete-continuous bi-level optimization problem. 
The goal is to select a subset $S_j$ of size $k$ from the training data $S \sim \mathbf{Z}^N$, such that a model trained on $S_j$ generalizes well to a held-out validation set $Q$. The bi-level objective can be expressed as:
$$\argmax_{S_j \subseteq S, |S_j| \leq k} \text{LL}_Q\big(\argmax_{\theta^T} \text{LL}_S(\theta^T_{S_j})\big)$$

where $\text{LL}_S$ and $\text{LL}_Q$ denote the training and validation log-likelihood functions, respectively.

The computational overhead in $\mathtt{Glister}$ arises from the subset selection. 
The outer optimization selects a subset $S_j$ from the training set using greedy or stochastic greedy algorithms. 
For a training set of size $N$, the naive greedy algorithm has a worst-case complexity of $O(N \cdot k \cdot Q \cdot f \cdot T / \gamma)$, where $k$ is the subset size and $\gamma$ is the frequency of subset selection. 
Stochastic greedy methods improve this to $O(N \cdot Q \cdot f \cdot T / \gamma \cdot \log (1/\epsilon))$, where $\epsilon$ is the tolerance level. 
Thus, the total computational overhead (in the best case with stochastic greedy optimization) for subset selection can be expressed as :
$$\boxed{\text{Computational Overhead}_\mathtt{Glister} = O\Big(\frac{NQTf}{\gamma}\log (1/\epsilon)\Big)}$$
The storage overhead for $\mathtt{Glister}$ includes storing loss values of the $Q$ validation set samples in addition to the regular training, resulting in,
$$\boxed{\text{Storage Overhead}_\mathtt{Glister} = O(Q)}$$

\paragraph{Example Forgetting ($\mathtt{Forgetting}$) \cite{forgetting_toneva2018}} 
This metric measures the number of times individual training samples transition from being correctly classified to incorrectly classified during training. This provides a way to rank examples based on their learning dynamics, with frequently forgotten examples considered more critical to the model's learning process.

The computational overhead of $\mathtt{Forgetting}$ arises from the need to track classification performance for every training sample throughout the training process. 
Specifically, for a dataset $S \sim \mathbf{Z}^N$ with $N$ training samples, $\mathtt{Forgetting}$ events are tracked over $T$ epochs of training. 
Since the model's prediction of all samples are readily available during training, no additional compute overhead is added. 
$$\boxed{\text{Computational Overhead}_{\mathtt{Forgetting}} = 0}$$
The storage overhead of $\mathtt{Forgetting}$ includes storing the classification results or loss values for each training sample across all epochs resulting in
$$\boxed{\text{Storage Overhead}_\mathtt{Forgetting} = O(NT)}$$

\paragraph{GraphCut-based Data Subset Selection ($\mathtt{GraphCut}$) \cite{graphcut_iyer2021}} 
The $\mathtt{GraphCut}$ method selects representative subsets of data by leveraging a submodular objective based on a generalized graph cut function. The objective is defined as:
$$ f(S_j) = \lambda \sum_{m \in S} \sum_{j \in S_j} s(m, j) - \sum_{j_1, j_2 \in S_j} s(j_1, j_2)$$
where $s(a_1, a_2)$ is the similarity kernel between data points, $\lambda \geq 2$ ensures monotonicity and submodularity, and $S$ is the ground set of all data points. 
The subset $S_j$ is selected to maximize the mutual information between $S_j$ and its complement $ S \setminus S_j$ under a cardinality constraint.

The computational overhead of $\mathtt{GraphCut}$ with a naive greedy approach arises from evaluating the marginal gain $f(j \mid S_j)$ for each element $j$ in the candidate set $S \setminus S_j$ at every iteration. 
For a dataset of size $N$, the marginal gain computation requires evaluating the similarity kernel $s(m, j) \forall m \in S, j \in S_j$ as well as $s(j_1, j_2) \forall j_1, j_2 \in S_j$. 
Since the subset $S_j$ has at most $k$ elements, the cost per gain evaluation is approximately $O(N \cdot k + k^2)$. Over $(N - k)$ iterations of subset selection, the total computational overhead is $O((N \cdot k + k^2)\cdot (N - k)) = N^2k - k^3$. Since $N > k$, we can express the total computational overhead in Big-$O$ notation as,
$$\boxed{\text{Computational Overhead}_\mathtt{GraphCut} = O(N^2k)}$$

The storage overhead includes storing the similarity kernel $s(i, j)$ for $N$ data points, which requires $O(N^2)$ memory. In addition, the subset $S_j$ being constructed requires storage for $k$ elements, and marginal gains for $N$ elements are stored at each iteration, resulting in an additional storage cost of $O(k + N)$. Thus, the total storage overhead is dominated by the similarity kernel and is given by,
$$\boxed{\text{Storage Overhead}_\mathtt{GraphCut} = O(N^2)}$$

\paragraph{Contrastive Active Learning ($\mathtt{Cal}$) \cite{cal_margatina2021}} 
The $\mathtt{Cal}$ method selects unlabeled examples for annotation by identifying contrastive examples in the pool of unlabeled data. 
These are data points that are similar to labeled examples in the feature space but have maximally different predictive likelihoods. 
The contrastive acquisition function is defined by a two-step process: (1) finding neighbors in the feature space using a similarity measure, and (2) ranking candidates by their divergence in predictive probabilities.

The computational overhead of $\mathtt{Cal}$ arises from two main components: nearest-neighbor searches and divergence computations. Given a labeled set $S$ of size $N$ and an query pool of size $Q$, nearest-neighbor searches involve computing pairwise distances between the feature representations of all $Q$ unlabeled data points and all $N$ labeled data points. Using a feature dimensionality $d$, this requires $O(N \cdot Q \cdot d)$ operations for each acquisition iteration. 

The second step computes the Kullback-Leibler (KL) divergence \cite{kullback1951} between the model’s predictive probability distributions for each unlabeled data point and its neighbors. Assuming $k$ nearest neighbors are considered, this step requires $O(Q \cdot k \cdot c)$, where $c$ is the number of classes. Since $NQd >> Qkc$, in big-$O$ notation,
$$\boxed{\text{Computational Overhead}_\mathtt{Cal} = O(NQd)}$$

The storage overhead for $\mathtt{Cal}$ includes storing the feature representations for all labeled and unlabeled data points, $(N + Q) \cdot d$. Additionally, the KL divergence scores for all $Q$ unlabeled candidates are stored during ranking, contributing an additional $Q$. Since $Nd >> Qd + Q$, in big-$O$ notation,
$$\boxed{\text{Storage Overhead}_\mathtt{Cal} = O(Nd)}$$.

\paragraph{Gradient Norm-based Data Pruning ($\mathtt{GraNd}$) \cite{E2LN_grand_paul2021}} 
The $\mathtt{GraNd}$ method identifies important examples early in training by ranking them based on their gradient norms. 
The score for a training sample $\Vec{z}_m$ at time $t$ is defined as the expected norm of the gradient of the loss function with respect to the model parameters, denoted as $\mathbb{E}_{\theta_t}\|\nabla_{\theta_t} \ell(\theta^{t}_S, \vec{z}_m)\|^2$. 
$\mathtt{GraNd}$ uses these scores to rank and prune unimportant training examples early in training, allowing models to train efficiently on smaller subsets of data.

The computational overhead of $\mathtt{GraNd}$ arises from two main components: gradient computation and score averaging. 
For a dataset $S$ with $N$ training samples and $p$ model parameters, the gradient norm for each sample must be computed at each iteration. 
Assuming $f$ represents the cost of a single forward pass and $2f$ represents the cost of a backward pass, the cost of computing the gradient norm for all samples in one epoch is approximately $N \cdot 3f$. 
To improve robustness, $\mathtt{GraNd}$ scores are averaged over multiple model initializations, denoted by $R$. Thus, for computing the scores over $T$ epochs and $R$ initialization,
$$\boxed{\text{Computational Overhead}_\mathtt{GraNd} = O(3NTRf)}$$

The storage overhead of $\mathtt{GraNd}$ includes storing gradient norms (of $p$ parameters) for all $N$ training samples across $T$ epochs and $R$ initializations. Thus,
$$\boxed{\text{Storage Overhead}_\mathtt{GraNd} = O(NTRp)}$$

\paragraph{Kernel Herding ($\mathtt{Herding}$) \cite{herding_chen2010}} selects representative samples by iteratively constructing a subset that approximates the true distribution of a dataset. 
Using a kernel-based approach, $\mathtt{Herding}$ generates a sequence of samples that greedily minimize the error in approximating the distribution in a \textit{Reproducing Kernel Hilbert Space} (RKHS). 
At each iteration $t$, the sample $\vec{z}_m$ is selected to maximize:
$$\vec{z}^{*t}_m = \argmax_{\vec{z}_m \in S} \langle w_{t-1}, \phi(\vec{z}_m) \rangle$$
where $\phi(\vec{z}_m)$ is the feature map for the kernel, $w_{t-1}$ is the weight vector at iteration $t-1$, and $S$ is the dataset.

The computational overhead of $\mathtt{Herding}$ arises primarily from evaluating the kernel similarity between the current candidate $\vec{z}_m \in S\setminus S_j$ and all previously selected samples $S_j = \{\vec{z}^{*t'}_m , \forall t' < t\}$ . For a dataset $S$ with $N$ samples, the cost of computing the similarity kernel for all $N$ data points at each of the $T$ iterations is $O(N \cdot T \cdot d)$, where $d$ is the dimensionality of the feature space. Additionally, updating the weight vector $w_t$ at each step requires $O(T \cdot d)$ operations. Thus,
$$\boxed{\text{Computational Overhead}_\mathtt{Herding}=O(NTd)}$$

The storage overhead for $\mathtt{Herding}$ includes storing the similarity kernel for all pairs of selected samples, which scales as $O(T^2)$. Additionally, the feature map for all samples in the dataset must be stored, requiring $O(N \cdot d)$ memory. Since $Nd >> T^2$ in big-$O$ notation, 
$$\boxed{\text{Storage Overhead}_\mathtt{Herding}=O(Nd)}$$

\paragraph{Using Curvature ($\mathtt{Slocurv}$) \cite{slocurves_garg2023}} It computes the input curvature $\mathtt{Curv}$, a proxy TDA for each training sample at the end of training as
\begin{equation*}
    \dfrac{1}{R}\sum_{R} \left\Vert \dfrac{\partial\left( \ell(\theta^{T}_S, [\vec{z}_m +hv]) - \ell(\theta^{T}_S, [\vec{z}_m]) \right)}{\partial \vec{z}_m} \right\Vert^2_2
\end{equation*}
Here, the hyperparameters $R$ represent the number of ``repeats" (\textit{not model retrainings}) done to get an empirical expectation of the randomly generated $hv$ which represents the Rademacher random variables used in Hutchinson's trace estimator \citep{hutchinson1989stochastic}.
Then the samples with the lowest $\mathtt{Curv}$ are chosen to form the coreset. 

For each training sample ($N$), and each repeat $R$,  2 forward passes and 1 backward pass are required. Assuming that $\ell(\theta^{T}_S, [\vec{z}_m])$ can be estimated only once, this requires a total of $R+1$ forward and $R+1$backward passes for each sample per epoch. Hence,
$$\boxed{\text{Computational Overhead}_\mathtt{Slocurv}=O(3NRf)}$$

The storage overhead for $\mathtt{Slocurv}$ arises primarily from storing gradients and random perturbations used during the computation of the curvature for each training sample. 
The main memory requirement comes from storing the gradients $\frac{\partial \ell}{\partial \vec{z}_m}$ for all samples $\vec{z}_m \in S$. 
For a dataset of size $N$ and input dimensionality $d$, the gradients must be stored across $R$ repeats, leading to a storage cost of $O(R \cdot N \cdot d)$.
Additionally, the Rademacher random perturbations $hv$ used in Hutchinson’s trace estimator must be stored for all $R$ repeats, adding an additional $O(R \cdot d)$.
Intermediate loss values $\ell(\theta^{T}_S, [\vec{z}_m])$ and their perturbed counterparts $\ell(\theta^{T}_S, [\vec{z}_m + hv])$ are computed and stored temporarily during forward and backward passes, but these contribute negligible additional storage compared to the gradient and perturbation storage. Since $NRd >> Rd$,
$$\boxed{\text{Storage Overhead}_\mathtt{Slocurv}=O(NRd)}$$

\textbf{Loss Trajectory Correlations ($\mathtt{LTC}$)} Unlike previous coreset methodologies, we only need access to the loss values of the training ($N$) and query samples ($Q$) throughout training. Since loss values of the training samples are directly available during training, we only need to perform forward passes (with $f$ FLOPs) for each epoch $T$, for all the query samples $Q$, resulting in 
$$\boxed{\text{Computational Overhead}_\mathtt{LTC} = O(QTf)}$$

In terms of storage, we need to store the loss values of all the training and query samples across the epoch, since $N>>Q$, 
$$\boxed{\text{Storage Overhead}_\mathtt{LTC} = O(NT)}$$

\subsection{Summary and Example Setting}
\label{appendix:summarized_coreset_overheads}
To further illustrate the computational efficiency of $\mathtt{LTC}$, we provide approximate overhead values by substituting the values of the parameters for an example using the ImageNet-1k dataset ($N=1,281,167$, $Q=50,000$, $d=224\times224\times3$, $c=1000$) and a ResNet-18 model with $p=11,689,128$ parameters and $f=1,818,228,160$ FLOPS per forward pass. 
We considered the training recipe provided by the popular Bearpaw library \citep{bearpaw_github}, where $T=90$.
For the $\mathtt{Slocurv}$, $\mathtt{GraNd}$ methods we used $R=10$, and for $\mathtt{Glister}$ we used $\gamma=1$. $\epsilon = 0.01$ as recommended by the authors.
The results are summarized in \cref{tab:appendix_coresets_overheads}.
We assume each parameter requires 4B of disk space and express the computational overhead in terms of \textit{peta ($10^{15}$) floating-point operations per second} (PFLOPs).

\begin{table*}[ht]
\centering
\caption{Summarized overheads of coreset methodologies along with a quantitative illustration to demonstrate the orders of magnitude of overhead of each technique. }
\vspace{2mm}
\label{tab:appendix_coresets_overheads}
\renewcommand{\arraystretch}{1.5}
\begin{tabular}{|c|c|c|r|r|}
\hline
\rowcolor[HTML]{EFEFEF} 
\cellcolor[HTML]{EFEFEF} &
  \cellcolor[HTML]{EFEFEF} &
  \cellcolor[HTML]{EFEFEF} &
  \multicolumn{2}{c|}{\cellcolor[HTML]{EFEFEF}\textbf{ImageNet-1k, ResNet-18}} \\ \cline{4-5} 
\rowcolor[HTML]{EFEFEF} 
\multirow{-2}{*}{\cellcolor[HTML]{EFEFEF}\textbf{Method}} &
  \multirow{-2}{*}{\cellcolor[HTML]{EFEFEF}\textbf{\begin{tabular}[c]{@{}c@{}}Computational\\ Overhead\end{tabular}}} &
  \multirow{-2}{*}{\cellcolor[HTML]{EFEFEF}\textbf{\begin{tabular}[c]{@{}c@{}}Storage\\ Overhead\end{tabular}}} &
  \multicolumn{1}{c|}{\cellcolor[HTML]{EFEFEF}\textbf{Comp. Overhead}} &
  \textbf{Sto. Overhead} \\ \hline \hline
$\mathtt{Glister}$ & $O(\frac{NQTf}{\gamma}\log (1/\epsilon))$ & $O(Q)$ & $\approx 2.10  \times 10^{7}$ PFLOPs & $\approx 2 \times 10^{-4}$ GB\\\hdashline
$\mathtt{Forgetting}$ & - & $O(NT)$ & - & $\approx 0.4$ GB \\\hdashline
$\mathtt{GraphCut}$ & $O(N^2k)$ & $O(N^2)$ & $\approx 2.10 \times 10^{2}$ PFLOPs & $\approx 6.56 \times 10^{3}$ GB \\\hdashline
$\mathtt{Cal}$ & $O(NQd)$ & $O(Nd)$ & $\approx 9.64$ PFLOPs & $\approx 7.71 \times 10^2$ GB \\\hdashline
$\mathtt{GraNd}$ & $O(3NTRf)$ & $O(NTRp)$ & $\approx 6.29 \times 10^3$ PFLOPs & $\approx 5.39 \times 10^7$ GB \\\hdashline
$\mathtt{Herding}$ & $O(NTd)$ & $O(Nd)$ & $\approx 1.74 \times 10^{-2}$ PFLOPs & $\approx 7.71 \times 10^2$ GB\\\hdashline
$\mathtt{Slocurv}$ & $O(3NRf)$ & $O(NRd)$ & $\approx 69.9$ PFLOPs & $\approx 7.71 \times 10^3$ GB\\\hdashline
\rowcolor[HTML]{E6E6FF} $\mathtt{LTC}$ (Ours) & $O(QTf)$ & $O(NT)$ & $\approx 8.18$ PFLOPs & $\approx 0.4$ GB\\\hline

\end{tabular}
\end{table*}

\section{Training Data Attribution (TDA) Methods}
\label{appendix:TDA_metrics_overview}
The earliest TDA methods utilized \textit{Leave-One-Out} (LOO) training, which involves retraining the model after removing specific data points and observing the changes in performance. 
While straightforward, LOO retraining is computationally prohibitive for modern deep learning models due to the need for multiple retraining cycles~\cite{Koh2017}. 
Recent TDA metrics, such as FZ-Influence ($\mathtt{Infl}$)~\cite{FZ_infl_feldman2020} and $\mathtt{Datamodels}$~\cite{datamodels_ilyas2022}, have gained popularity owing to precomputed scores for widely-used datasets in computer vision. 
These methods, however, face scalability challenges. 

A prominent alternative that arose was \textit{Influence Functions}, which estimated the effect of downweighting individual samples using first-order (gradient) and second-order (Hessian) computations~\cite{Koh2017, influence_fragile_basu2021} performed at the end of training. 
Methods like $\mathtt{RandSelect}$~\cite{Randselect_wojnowicz2016} and $\mathtt{Arnoldi}$ iterations~\cite{Arnoldi_schioppa2022} improved computational efficiency by approximating the Hessian. 
Similarly, $\mathtt{TRAK}$~\cite{trak_2023park} combined random projections, gradient-based methods, and ensembling to estimate the influence of training samples. 
However, these approaches often rely on strong assumptions, such as convergence to a unique optimal solution, which limits their applicability to neural networks. 
Additionally, Hessian computations introduce significant computational overhead.
To address these challenges, \textit{unrolling-based} methods that observe the learning process across training iterations have been proposed. 
These techniques approximate the impact of samples by differentiating through the optimization trajectory~\cite{dataclensing_hara2019}. 
Among these, $\mathtt{TracIn}$~\cite{tracin_pruthi2020} is a highly efficient method that estimates influence using gradients tracked throughout training. 
Its practical implementation, $\mathtt{TracInCP}$, uses intermediate checkpoints to alleviate computational burdens. 
While effective, unrolling methods require storing intermediate training states, leading to high storage and computational costs.

\subsection{Computational and Storage Overheads}
\label{appendix:detailed_tda_overheads}

\begin{table*}[ht]
\centering
\caption{Comparison of computational and storage overheads of $\mathtt{LTC}$ with existing TDA methods. The details of the notation are outlined in \cref{sec:coreset_overheads}.}
\vspace{1mm}
\label{tab:TDA_overheads}
\renewcommand{\arraystretch}{1.5}
\begin{tabular}{|c|c|c|}
\hline
\rowcolor[HTML]{D8D8D8} \textbf{Method} & \textbf{Computational Overhead} & \textbf{Storage Overhead} \\\hline \hline
$\mathtt{Infl}$ & $O(3\lceil\alpha N\rceil QRTf)$ & $O(Rp)$ \\
$\mathtt{Datamodels}$ & $O(3\lceil \alpha N\rceil RTf)$ & $O(RN)$ \\ 
$\mathtt{TRAK}$ & $O(2NRp \cdot p')$ & $O(Np')$ \\
$\mathtt{Arnoldi}$ & $O(2NQp \cdot p')$ & $O(NQp)$ \\ 
$\mathtt{TracIn}$ & $O(3NTf)$ & $O(NTp)$ \\\hdashline
\rowcolor[HTML]{E6E6FF} $\mathtt{LTC}$ (Ours) & $O(QTf)$ & $O(NT)$ \\ \hline
\end{tabular}
\end{table*}

$\mathtt{LTC}$ also has lower storage and computational overhead compared to TDA metrics that measure influence. 
Utilizing the same notation as used in \cref{sec:coreset_overheads}, we also compare the overheads of TDA metrics with $\mathtt{LTC}$. 
This is shown in \cref{tab:TDA_overheads}. 
Here $R$ is the number of model retrainings required and $p'$ is a lower dimension ($p' \ll p$) used to alleviate computation. 
$\alpha$ is a fractional value $\in (0, 1]$ and $b$ is the batch size used to increase computation speed.  
$\mathtt{LTC}$ exhibits significantly lower computational complexity compared to these metrics. 
The details are outlined below. 
\subsubsection{LOO Methods}
\paragraph{FZ-Influence ($\mathtt{Infl}$) \cite{FZ_infl_feldman2020}} The influence score $\mathtt{Infl}$ of a training sample $\vec{z}_m$ on a query sample $\vec{z}_q$ is given by:
\begin{align*}
\mathtt{Infl}(\vec{z}_q, \vec{z}_m, S; \lambda) \coloneqq \mathop{\mathbb{E}}_{\xi} \big[ & \Pr(\theta^T_{S}(\vec{x}_q) = \vec{y}_q) \notag \\
& - \Pr(\theta^T_{S\setminus m}(\vec{x}_q) = \vec{y}_q) \big]
\end{align*}

Here, $\Pr(\theta^T_{S}(\vec{x}_q) = \vec{y}_q)$ represents the confidence in the prediction for $\vec{z}_q$ using a model trained on $S$, and $S^{\setminus m}$ is the training set with sample $\vec{z}_m$ removed. 
This expression measures the change in the prediction probability for the query sample $\vec{z}_q$ when the training sample $\vec{z}_m$ is included in the training set compared to when it is excluded. 
The underlying intuition is that if sample $\vec{z}_m$ significantly influences the prediction for $\vec{z}_q$, its exclusion will lead to a notable reduction in prediction accuracy for $\vec{z}_q$, resulting in a higher influence score.
This metric ideally require retraining $N+1$ models to calculate the influence scores of all pairs of samples. 
To reduce the retraining costs, the authors of the metrics suggest training a smaller number ($R$) of models with a fraction ($\alpha$) of the training dataset ($|S_j| = \lceil \alpha |S| \rceil$). 
Each of these subsets is selected randomly. 
They then estimate $\Pr(\theta^T_{S}(\vec{x}_q) = \vec{y}_q)$ to be the average prediction probability of  $\vec{z}_q$ of all models that were trained with a subset $S_j$ that included the sample $m$ ($m \in S_j$), and $\Pr(\theta^T_{S\setminus m}(\vec{x}_q) = \vec{y}_q)$ to be the average prediction probability of  $\vec{z}_q$ of all models that were trained with a subset $S_{j'}$ that excluded the sample $m$ ($m \notin S_{j'}$). 

Hence the total computational overhead would be to train $R$ models for $T$ epochs (run forward and backward passes for each sample $RT$ times ), and for each model, we need to calculate the prediction probability (or run a forward pass) for $\lceil\alpha N\rceil$ train samples and $Q$ test samples. Thus,
$$\boxed{\text{Computational Overhead}_\mathtt{Infl} = O(3\lceil\alpha N\rceil QRTf)}$$
In addition, there would be a need to store the $R$ model parameters for influence score calculation, resulting 
$$\boxed{\text{Storage Overhead}_\mathtt{Infl} = O(Rp)}$$

\paragraph{Datamodel Metric \cite{datamodels_ilyas2022}}
For a fixed query sample $\vec{z}_q$, $\mathtt{Datamodels}$ is a parametric function $g_\theta$ trained to predict the outcome of a model $\theta^T_{S_j}$ trained on a subset $S_j$ of the training data $S$ on $\vec{z}_q$. 
The $\mathtt{Datamodels}$ function $g_\theta$ is typically instantiated as a linear function of the characteristic vector of $S_j$, allowing for more efficient computation and analysis.

The computational overhead of constructing $\mathtt{Datamodels}$ is determined by three steps: sampling subsets, training models on these subsets, and solving the regression problem to estimate the $\mathtt{Datamodels}$ parameters. 

To construct the $\mathtt{Datamodels}$, $R$ subsets $S_j$ of size $\lceil\alpha N\rceil$ are sampled from the training data. A model is then trained on each subset, and the output $\theta^T_{S_j}(\vec{z}_q)$ is computed. This incurs a cost of $(R \cdot T \cdot \alpha N \cdot 3f)$ where $3f$ accounts for the forward and backward passes required during training.

The regression step fits the $\mathtt{Datamodels}$ function $g_\theta$ by solving an $l_1$-regularized least squares problem over $R$ samples, where the feature vector dimension corresponds to the training set size $N$. Using efficient solvers optimized for sparse regression, the computational cost of this step scales as $O(NR)$.
Since $3\lceil \alpha N\rceil RTf > NR$, using big-$O$ notation,
$$\boxed{\text{Computational Overhead}_\mathtt{Datamodels} = O(3\lceil \alpha N\rceil RTf)}$$

The storage overhead includes storing the $NQ$ dimensional $\mathtt{Datamodels}$ training matrix, which consists of $R$ characteristic vectors of size $Q$, and storing the outputs $\theta^T_{S_j}$ for each subset. Since $N>>Q$,
$$\boxed{\text{Storage Overhead}_\mathtt{Datamodels} = O(RN)}$$

\subsubsection{Implicit Differentiation TDA Techniques}
\paragraph{Tracing with the Randomly-projected After Kernel $\mathtt{TRAK}$ Score \cite{trak_2023park}} It is a scalable data attribution method that leverages linearization via the empirical neural tangent kernel (eNTK) and random projections to approximate model behavior. 
It achieves data attribution by training $R$ models, each on a subset of the training data, and computing influence scores using gradient features projected to a lower-dimensional space.

The total computational overhead for $\mathtt{TRAK}$ is composed of three main components: model training, gradient computation, and random projections. 
For model training, $R$ models are trained on subsets of the training set of size $\lceil\alpha N\rceil$, where $\alpha$ is the fraction of the training set used per subset. 
Each model is trained for $T$ epochs, with the cost of a forward and backward pass for each sample being $3f$ FLOPs. Since $R-1$ additional models need to be trained,
the additional training cost is $3\lceil\alpha N\rceil (R-1)Tf$.
Next, for each of the $R$ trained models, gradients are computed for all $N$ training samples and $Q$ test samples. 
Each gradient computation involves one forward and one backward pass, resulting in a cost of $3R(N + Q)3f$. 
The gradient features are then projected to a lower dimension $p'$ using a random projection matrix. 
Each projection operation for a single sample requires the computation of a matrix-vector product of size $p \times p'$, which results in $2p \cdot p'$ FLOPs (accounting for one multiplication and one addition), leading to an additional cost of $2R(N + Q)p \cdot p'$. 
Thus, in big-$O$ notation, 
$$\boxed{\text{Compuational Overhead}_\mathtt{TRAK}=2NRp\cdot p'}$$

The storage overhead for $\mathtt{TRAK}$ consists of two main components. 
First, the $R$ trained models need to be stored, resulting in a storage cost of $R \cdot p$. 
Second, the gradients of $N$ training samples and $Q$ test samples are projected into a space of dimension $p'$. This adds a storage requirement of $(N + Q) \cdot p'$. Thus, 
$$\boxed{\text{Storage Overhead}_\mathtt{TRAK}=O(Np')}$$

\paragraph{$\mathtt{Arnoldi}$ Iterations \cite{Arnoldi_schioppa2022}} 
This method computes influence scores by approximating the inverse Hessian using $\mathtt{Arnoldi}$ iteration. 
By diagonalizing the Hessian $H$ in a low-dimensional subspace spanned by the top eigenvectors, it aims to alleviate storage and computational overhead. 
The influence score between a training example $\vec{z}_m$ and a query example $\vec{z}_q$ is calculated as:
\begin{align*}
\mathtt{Arnoldi}(\vec{z}_q, \vec{z}_m, S, \lambda) = \langle  \nabla_\theta \ell(\theta^T_S,\vec{z}_q), H^{-1}  \nabla_\theta \ell(\theta^T_S,\vec{z}_m) \rangle   
\end{align*}
where $H =  \nabla^2_\theta \ell(\theta^t_S,\vec{z})$ is the Hessian of the loss $\ell$ with respect to the model parameters $\theta$.

The computational overhead is primarily composed of three steps: \textit{Hessian-vector products} (HVPs), $\mathtt{Arnoldi}$ iteration to diagonalize $H$, and projections to compute influence scores. 

Each HVP computes the product $H \cdot \vec{v}$ for a vector $\vec{v}$ and can be efficiently implemented as a combination of reverse- and forward-mode differentiation. 
For a model with $p$ parameters, the HVP computation involves backpropagating through the network for a batch of size $b$. 
The cost for a single HVP is therefore $3bpf$.
Here, the factor of $3f$ accounts for one forward and one backward pass per HVP.
$\mathtt{Arnoldi}$ iteration computes the dominant eigenvectors of $H$ over $R$ iterations. 
The total cost of diagonalization, accounting for $R$ HVPs, is $(3Rbpf)$.

Once $H$ is diagonalized in a subspace of dimension $p' \ll p$, influence scores are computed using projections. 
Each projection involves a matrix-vector multiplication requiring $2p \cdot p'$ FLOPs (accounting for one multiplication and one addition). 
For $NQ$ training-query pairs, the total cost of projections is $2NQp \cdot p'$. 
Since $2NQp\cdot p' >> 3Rbpf$,
$$\boxed{\text{Computational Overhead}_\mathtt{Arnoldi}=O(2NQp\cdot p')}$$

The storage overhead consists of storing the reduced subspace and gradients. 
The subspace requires $p' \cdot p$ storage, while gradients for $NQ$ samples require $NQp$ storage. Since the gradients are larger, in big-$O$ notation,
$$\boxed{\text{Storage Overhead}_\mathtt{Anroldi}=O(NQp)}$$

\subsubsection{Unrolled TDA Techniques}
\paragraph{$\mathtt{TracIn}$ Score \cite{tracin_pruthi2020}}
It computes the influence as:
\begin{align*}
\mathtt{TracIn}(\vec{z}_q, \vec{z}_m, S; \lambda) \coloneqq \sum_{t=0}^T \nabla_\theta \ell(\theta^t_S,\vec{z}_m) \cdot \nabla_\theta \ell(\theta^t_S,\vec{z}_q)
\end{align*}
This metric requires computing the gradient (one forward and one backward pass) for each sample in the train set of size $N$, and the query set of size $Q$ for each epoch ($T$ times). Since $N>>Q$,
$$\boxed{\text{Computational Overhead}_\mathtt{TracIn}=O(3NTf)}$$

The gradients of each of the samples need to be trained for each iteration, resulting in 
$$\boxed{\text{Storage Overhead}_\mathtt{TracIn}=O(NTp)}$$

\end{document}